\pdfoutput=1
\documentclass[10pt,twocolumn,letterpaper]{article}

\usepackage[pagenumbers]{cvpr}

\usepackage{microtype}
\usepackage{xcolor}
\usepackage{xspace}
\usepackage{amsthm}
\usepackage{needspace}
\usepackage{algorithm}
\usepackage[noend]{algpseudocode}

\renewcommand{\paragraph}[1]{\vspace{.30em}\noindent\textbf{#1}\hspace{0.5em}}

\apptocmd{\normalsize}{%
  \setlength{\abovedisplayskip}{5pt plus 2pt minus 2pt}%
  \setlength{\belowdisplayskip}{5pt plus 2pt minus 2pt}%
  \setlength{\abovedisplayshortskip}{0pt plus 2pt}%
  \setlength{\belowdisplayshortskip}{4pt plus 2pt minus 2pt}%
}{}{}

\newcommand{\methodname}{Rattan\xspace}
\newcommand{\nexus}{Nexus\textsuperscript{$\dagger$}\xspace}
\newcommand{\corr}{{\fontfamily{mvs}\fontencoding{U}\selectfont\char66}}

\newcommand\blfootnote[1]{%
  \begingroup
  \renewcommand\thefootnote{}\footnote{#1}%
  \addtocounter{footnote}{-1}%
  \endgroup
}

\algrenewcommand\algorithmicindent{0.9em}
\makeatletter
\AtBeginDocument{\renewcommand{\theHALG@line}{\thealgorithm.\arabic{ALG@line}}}
\patchcmd{\@maketitle}{\vskip .375in}{\vskip .16in}{}{}
\patchcmd{\@maketitle}{\vspace*{24pt}}{\vspace*{10pt}}{}{}
\patchcmd{\@maketitle}{\vspace*{12pt}}{\vspace*{0pt}}{}{}
\makeatother

\newtheoremstyle{methodplain}{4pt}{4pt}{\itshape}{}{\bfseries}{.}{0.45em}{}
\newtheoremstyle{methodremark}{3pt}{3pt}{\normalfont}{}{\itshape}{.}{0.45em}{}
\theoremstyle{methodplain}

\newtheorem{proposition}{Proposition}
\theoremstyle{methodremark}

\definecolor{cvprblue}{rgb}{0.21,0.49,0.74}
\usepackage[pagebackref,breaklinks,colorlinks,allcolors=cvprblue]{hyperref}
\hypersetup{
  pdftitle={Rattan: Growing Meshes with Structured Discrete Flows},
  pdfauthor={Dehao Hao, Kaiyi Zhang, Haohan Weng, Junliang Ye, Xiangjun Gao, Zibo Zhao, Zhuo Chen, Chunchao Guo, Long Quan},
}

\title{\methodname: Growing Meshes with Structured Discrete Flows}

\author{%
Dehao Hao\textsuperscript{1\,$*$}\quad
Kaiyi Zhang\textsuperscript{1\,$*$}\quad
Haohan Weng\textsuperscript{2}\quad
Junliang Ye\textsuperscript{2}\quad
Xiangjun Gao\textsuperscript{1}\\[2pt]
Zibo Zhao\textsuperscript{2}\quad
Zhuo Chen\textsuperscript{2}\quad
Chunchao Guo\textsuperscript{2\,\corr}\quad
Long Quan\textsuperscript{1\,\corr}\\[5pt]
\textsuperscript{1}\,The Hong Kong University of Science and Technology\qquad
\textsuperscript{2}\,Tencent Hunyuan
}

\begin{document}
\twocolumn[{%
  \renewcommand\twocolumn[1][]{#1}%
  \vspace{-12pt}
  \maketitle
  \vspace{-18pt}
  \begin{center}
    \captionsetup{type=figure}
    \includegraphics[width=0.93\textwidth]{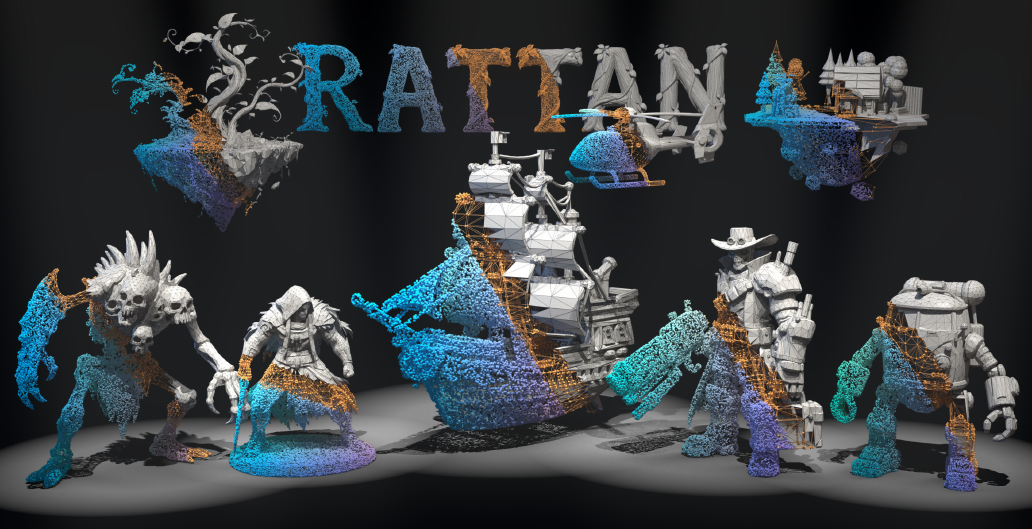}
    \captionof{figure}{\textbf{\methodname} generates compact, high-fidelity artistic meshes across characters, vehicles, and scenes by growing vertices and edges via structured discrete flows.}
    \label{fig:teaser}
    \vspace{-4pt}
  \end{center}%
}]
\blfootnote{\textsuperscript{$*$}\,Equal contribution.\quad \textsuperscript{\corr}\,Corresponding authors.}

\begin{abstract}

Generative modeling of unstructured meshes requires learning a distribution over vertex sets of varying cardinality and their combinatorial connectivity. Existing approaches typically transform meshes into alternative representations: autoregressive models serialize meshes into sequences, imposing an artificial ordering, while latent diffusion models rely on specialized autoencoders to map discrete mesh structures into continuous latent spaces. In contrast, we introduce \textbf{\methodname}: a unified continuous-time Markov chain (CTMC) framework that generates meshes from scratch through progressive discrete structural transitions. This supports native, parallelizable mesh generation without serialization or learned latents. Specifically, \textbf{Octree Flow} expands an octree across depths to adaptively resolve variable-cardinality vertices, while \textbf{Edge Flow} progressively inserts edges via vertex-centered proposals to establish sparse connectivity.
On point-cloud-conditioned benchmarks, Octree Flow and Edge Flow each surpass prior state-of-the-art counterparts, and together \methodname outperforms existing methods by a large margin, including when converting generated dense meshes into compact artistic meshes, while converging up to an order of magnitude faster in training across vertex and connectivity generation.
\methodname further generalizes to single-image-to-mesh generation with better geometric and topological quality.
More broadly, our formulation offers a general recipe for variable-cardinality structures grown through discrete transitions.

\end{abstract}

\section{Introduction}
\label{sec:intro}

Meshes are the foundational representation of 3D assets in graphics and digital content creation. Recent large-scale 3D generative models~\citep{zhang2024clay,li2025triposg,hunyuan3d2025hunyuan3d21,xiang2025trellis,xiang2025trellis2} achieve remarkable geometric fidelity by modeling continuous neural fields or volumetric grids and extracting surfaces via grid contouring~\citep{lorensen1987marching,ju2002dual,Nielson2004DualMC}. While capturing fine geometry, isosurface extraction yields over-tessellated, grid-induced triangulations, whereas rendering, animation, and editing call for compact artistic meshes with clean vertex and face layouts.

Artistic mesh generation addresses this need by modeling variable-cardinality vertex sets $V$ and sparse connectivity $E$ directly, preserving sharp features and compact polygons. Dominant autoregressive methods~\citep{nash2020polygen,siddiqui2024meshgpt,chen2025meshanything,hao2024meshtron} flatten meshes into 1D token sequences; despite compact tokenizations~\citep{tang2025edgerunner,weng2025bpt,lionar2025treemeshgpt,lin2026meshripple} and post-training~\citep{zhao2025deepmesh}, serialization imposes an artificial canonical order on an unordered 3D graph, causing slow sequential decoding and error drift. For parallel generation, non-autoregressive methods turn to diffusion and flow matching, from denoising unindexed triangle soups~\citep{alliegro2023polydiff,he2025meshcraft} to generating indexed meshes via continuous latents~\citep{li2026meshflow,zhao2026lato,wang2026polyflow,xu2026meshyt2}, with recent works~\citep{wang2026nexus,long2026lato2} converging on $p(V\mid c)\,p(E\mid V,c)$ under condition $c$ to decouple vertex placement from wiring.

Yet even under this factorization, neither factor is generated by a single process directly on its discrete structure. For $V$, autoregressive generators~\citep{nash2020polygen,kim2025fastmesh} still prescribe a token order, while parallel methods rely on depth-wise cascades over relaxed occupancies (Nexus~\citep{wang2026nexus}) or a vertex autoencoder on a voxel scaffold (LATO.2~\citep{long2026lato2}). For $E$, flow-based methods~\citep{wang2026polyflow,xu2026meshyt2,wang2026nexus,long2026lato2} encode discrete adjacency into continuous latents via an autoencoder, incurring a reconstruction--generation gap from lossy decoding and imperfect latent modeling.

Our key insight is that both factors admit \emph{monotone discrete growth} (Fig.~\ref{fig:pipeline}): expanding an octree from its root for $V$, and inserting edges among vertices for $E$. Building on discrete flow and generator matching~\citep{campbell2022continuous,gat2024discrete,holderrieth2025generator}, we formulate \emph{Growth Flows}, CTMCs decoupling feasible construction orders from continuous event timing to yield closed-form conditional rates and simulation-free training. \methodname instantiates two such flows directly on discrete mesh structures (Fig.~\ref{fig:teaser}). In \emph{Octree Flow}, a uniform prior over topological tree orderings induces exact per-node rates proportional to remaining subtree size, prioritizing detail-rich regions across depths in parallel via $\tau$-leaping~\citep{campbell2022continuous} without level-by-level barriers. In \emph{Edge Flow}, pure insertion evolves exclusively over the $O(|V|)$ target edges on a sparse graph, avoiding $O(|V|^2)$ dense pair states, while decomposing rates into per-vertex degree intensities and competitive neighbor pointers. Operating natively on octrees and sparse graphs, without serialization, learned latents, relaxations, or cascades, yields faster convergence, higher fidelity, single-image-to-mesh generation, and direct conversion of dense 3D generative outputs into compact, artist-ready meshes:
\begin{itemize}
    \item \textbf{Growth Flows}, CTMCs for monotone discrete structural growth with closed-form conditional rates for simulation-free generator matching~\citep{holderrieth2025generator}.
    \item \textbf{\methodname}, pairing subtree-guided cross-depth \emph{Octree Flow} for detail-adaptive vertex generation with sparse, vertex-centered \emph{Edge Flow} for parallel connectivity synthesis.
    \item \textbf{State-of-the-art quality and fast convergence:} Octree Flow and Edge Flow each surpass prior counterparts in factorized evaluations and together outperform prior point-cloud-conditioned methods by a large margin, converging ${4.9\times}\text{--}{12.0\times}$ faster than cascaded octree flow and ${7.3\times}\text{--}{10.3\times}$ faster than latent topology flow~\citep{wang2026nexus}.
\end{itemize}

\section{Related Work}
\label{sec:related}

\paragraph{3D Geometry Generation.}
Large-scale 3D generative models synthesize high-fidelity geometry via latents for continuous fields, as unstructured latent sets~\citep{zhang20233dshape2vecset,zhang2024clay,li2025triposg,hunyuan3d2025hunyuan3d21,Hao2026MSVSVAEMA} or sparse grids and octrees~\citep{ren2024xcube,xiong2025octfusion,wei2025octgpt,xiang2025trellis,xiang2025trellis2,li2026sparc3d,he2025sparseflex,wu2026direct3d}. Primal or dual contouring~\citep{lorensen1987marching,ju2002dual,shen2023flexicubes,xiang2025trellis2} extracts dense tessellations capturing fine geometry, complementing artistic mesh generation of compact vertex and face layouts.

\paragraph{Autoregressive Mesh Generation.}
PolyGen~\citep{nash2020polygen} generates vertices autoregressively followed by faces via a pointer network, whereas MeshGPT~\citep{siddiqui2024meshgpt} serializes faces into learned tokens. Follow-up works scale capacity and conditioning~\citep{chen2024meshxl,chen2025meshanything,hao2024meshtron}, design compact tokenizations~\citep{chen2025meshanythingv2,weng2025bpt,tang2025edgerunner,lionar2025treemeshgpt,song2026silksong,lin2026meshripple,xu2026sato}, apply reinforcement learning~\citep{zhao2025deepmesh,liu2025meshrft}, or decode edges in one step from autoregressive vertices~\citep{kim2025fastmesh}. Instead of prescribing a fixed token order, \methodname learns over random feasible growth orders and transitions in parallel.

\paragraph{Non-Autoregressive Mesh Generation.}
Triangle-soup generators~\citep{alliegro2023polydiff,he2025meshcraft,sun2026meshflow} denoise per-face coordinates or latents, leaving connectivity implicit in coincident vertices. Most other methods encode discrete adjacency into continuous latents or surface fields via a pretrained autoencoder~\citep{shen2024spacemesh,li2026meshflow,zhao2026lato,wang2026polyflow,xu2026meshyt2,li2026triflow,song2026meshbdf}, recovering edges upon decoding. Nexus~\citep{wang2026nexus} and LATO.2~\citep{long2026lato2} adopt this latent-topology paradigm after vertex generation: Nexus cascades per-depth flow matching on relaxed octree occupancies with a shared network, while LATO.2 flows structured latents on a voxel scaffold and decodes them into vertices. \methodname models each factor with one CTMC, growing octrees and edge sets directly without autoencoders, relaxations, or cascades.

\section{Methodology}
\label{sec:method}

\begin{figure*}[t]
    \centering
    \includegraphics[width=0.90\linewidth]{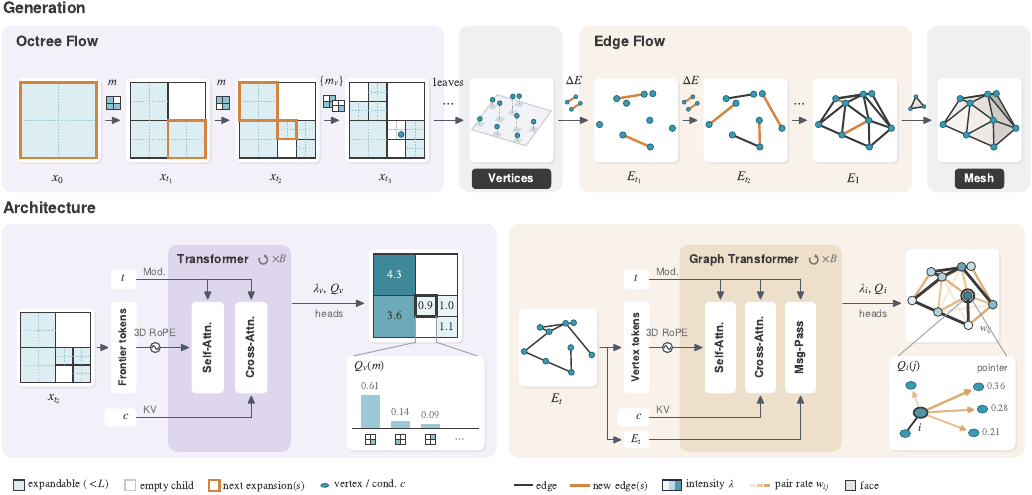}
    \caption{\textbf{Overview of \methodname.}
    \textbf{Top:} Octree Flow expands frontier nodes in parallel across depths to reveal child-occupancy patterns $m$ (2D quadtree, $L=3$, shown for clarity; amber: next expansions). Its leaves define the vertices, among which Edge Flow inserts edge batches $\Delta E$ (amber: new; dark: earlier) to form the mesh.
    \textbf{Bottom:} Conditioned on $c$, Transformers predict per-node intensities $\lambda_v$ and pattern distributions $Q_v(m)$ for Octree Flow, and per-vertex intensities $\lambda_i$ with neighbor pointers $Q_i(j)$ (pair rates $w_{ij}$ shown as opacity) for Edge Flow; darker fills indicate higher intensities.}
    \label{fig:pipeline}
\end{figure*}

\methodname factorizes mesh generation as
$p_\theta(V,E\mid c)=p_\theta(V\mid c)\,p_\theta(E\mid V,c)$
(Fig.~\ref{fig:pipeline}): \emph{Octree Flow} generates a
sparse octree indexing the vertices $V$, and \emph{Edge Flow}
generates their edge connectivity $E$ to form a triangle mesh.
Both stages are CTMCs that grow discrete structures from
scratch (Sec.~\ref{sec:growth}), trained by matching
conditional transition rates.

\subsection{Preliminaries}

\noindent\textbf{Continuous-time Markov Chains.}
A CTMC $\{X_t\}_{t\in[0,1]}$ on a discrete state
space $\mathcal X$ is specified by nonnegative
off-diagonal transition rates $r_t(x,y)$, with
$\Pr(X_{t+h}=y\mid X_t=x)=h\,r_t(x,y)+o(h)$ for $y\ne x$
(Appendix~\ref{app:ctmc}).

\noindent\textbf{Discrete Flow Matching.}
Discrete Flow Matching (DFM)~\citep{gat2024discrete}
constructs a probability path from a source
distribution $p_{\mathrm{src}}$ to the data
distribution $p_{\mathrm{data}}$ through
conditional CTMCs.
All distributions are conditioned on the input
condition $c$, which we leave implicit unless needed.
For $x_1\sim p_{\mathrm{data}}$, let
$p_t(x\mid x_1)$ be generated by
$r_t(x,y\mid x_1)$, with
$\mathbb E_{x_1}p_0(x\mid x_1)=p_{\mathrm{src}}(x)$
and $p_1(x\mid x_1)=\delta_{x_1}(x)$.
The marginal path
$p_t(x)=\mathbb E_{x_1}p_t(x\mid x_1)$
is generated by the posterior-averaged
rates~\citep{gat2024discrete,holderrieth2025generator}
\begin{equation}
    r_t(x,y)=\mathbb E
    \left[r_t(x,y\mid x_1)\mid X_t=x\right].
    \label{eq:pre_marginal}
\end{equation}
Without evaluating this posterior,
Generator Matching~\citep{holderrieth2025generator}
learns the marginal rates via
\begin{equation}
    \mathcal L_{\mathrm{CGM}}(\theta)
    =\mathbb E_{t,x_1,X_t\mid x_1}
    D_\phi\big(r_t(X_t,\cdot\mid x_1),\,r_t^\theta(X_t,\cdot)\big),
    \label{eq:pre_cgm}
\end{equation}
which equals marginal-rate matching up to a
$\theta$-independent constant under standard
conditions, so its minimizer is Eq.~\eqref{eq:pre_marginal}.
We use the generalized-KL Bregman divergence
over off-diagonal rates,
\begin{equation}
    D_\phi(a,b)=\sum_j
    \left[b_j-a_j+a_j\log\frac{a_j}{b_j}\right],
    \label{eq:pre_kl}
\end{equation}
with $0\log(0/b)=0$ for $b\ge 0$.

\subsection{Growth Flows}
\label{sec:growth}

\noindent\textbf{Growth Processes.}
Both stages are growth processes: states are partial
structures starting from a fixed initial state ($p_{\mathrm{src}}$
is a point mass) and reaching a target $x_1$ after $R$
events (expansions or edge insertions), each optionally
revealing a mark determined by $x_1$. Structural
dependencies constrain event order; we denote feasible
orders by $\mathcal O(x_1)$.

\noindent\textbf{Conditional Paths.}
Each flow uses a differentiable schedule $\kappa_t$
with $\kappa_0=0$, $\kappa_1=1$, and $\dot\kappa_t>0$
on $(0,1)$, defining hazard time
$\tau_t=-\log(1-\kappa_t)$ and hazard rate
$\alpha_t=d\tau_t/dt=\dot\kappa_t/(1-\kappa_t)$.
The conditional path separates event order from timing:
we draw a feasible order $\pi\in\mathcal O(x_1)$ from
a distribution specified by each flow and, independently,
$R$ i.i.d.\ timestamps with distribution function
$\kappa_t$, sorted and assigned to the events in order
$\pi$. At time $t$, the state comprises the first
$K_t$ events occurring by $t$.

\begin{proposition}[Conditional Growth Rates]
Given $x_1$,\label{prop:growth} the number $K_t$ of events by time $t$ is
$\mathrm{Binomial}(R,\kappa_t)$, independently of $\pi$.
If each $\pi_{k+1}$ depends on its prefix only through the
unordered set $A=\{\pi_1,\ldots,\pi_k\}$, then $X_t$ is a CTMC that,
from a state with revealed events $A$, $|A|=k<R$, reveals
event $v\notin A$, with its mark, at rate
\begin{equation}
    r_t\big(A,A\cup\{v\}\mid x_1\big)
    =\alpha_t(R-k)\Pr(\pi_{k+1}=v\mid A).
    \label{eq:growth_rate}
\end{equation}
\end{proposition}

Since $K_t$ is independent of $\pi$, training states
are sampled simulation-free via
$K_t\sim\mathrm{Binomial}(R,\kappa_t)$ and the
first $K_t$ events of $\pi$ (Appendix~\ref{app:growth}),
as satisfied by both flows below.

\noindent\textbf{Canonical Rates and Factorization.}
Writing $r_t=\alpha_t\bar r_t$ defines the
\emph{canonical rate} $\bar r_t$ per unit of $\tau_t$,
which both stages factorize into per-site intensities
$\lambda$ and categorical distributions $Q$. Since
$D_\phi(\alpha_t\bar a,\alpha_t\bar b)
=\alpha_t D_\phi(\bar a,\bar b)$ and positive time
weights preserve the minimizer of
Eq.~\eqref{eq:pre_cgm}~\citep{holderrieth2025generator},
we weight Eq.~\eqref{eq:pre_cgm} by $1/\alpha_t$ to
match canonical rates directly under
$t\sim\mathcal U(0,1)$, avoiding unbounded rate weights
and gradient variance as $t\to 1$.

\subsection{Octree Flow}
\label{sec:octree_flow}

\begin{figure}[t]
    \centering
    \includegraphics[width=0.92\columnwidth]{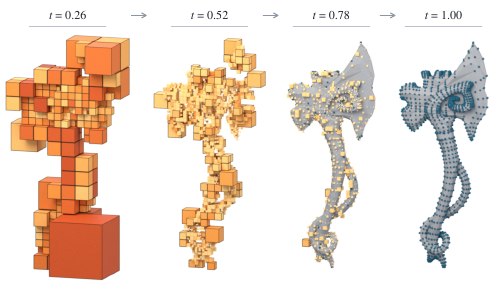}
    \caption{\textbf{Cross-depth Octree Flow trajectory} ($L=10$). Coarse cells, high-rate cells (warmer $\lambda_v$), and resolved leaf vertices (teal) coexist across depths to form $V$.}
    \label{fig:octree_growth}
\end{figure}

\noindent\textbf{Vertex Generation via Octree Expansion.}
We quantize mesh vertices onto a
$2^L\!\times2^L\!\times2^L$ grid
and represent occupied cells by a sparse
octree $T$ of depth $L$.
The centers of its depth-$L$ leaf cells give the vertices. Let
$\mathcal I(T)$ denote its internal nodes; the child
configuration of $v\in\mathcal I(T)$ is described by a
child-occupancy pattern
$m_v(T)\in\{1,\ldots,255\}$, the 8-bit mask of
its occupied children.

Octree Flow is a CTMC on the space $\mathcal X$ of
partial octrees (Fig.~\ref{fig:pipeline}, top left).
A partial octree $x\in\mathcal X$ contains the depth-$0$
root and all nodes created by previous expansions.
Let $\mathcal A(x)$ be its expanded internal
nodes, $\mathcal F(x)$ its unexpanded frontier
(including depth-$L$ leaves), and
$\mathcal S(x)=
\{v\in\mathcal F(x):
\operatorname{depth}(v)<L\}$
the expandable frontier.
The chain moves by \emph{expansions}: for
$v\in\mathcal S(x)$ and a pattern $m$, the state
$x\oplus(v,m)$ reveals $m$ at $v$ and instantiates its
occupied children $\mathrm{ch}_m(v)$,
\begin{equation}
    \begin{aligned}
    \mathcal A\big(x\oplus(v,m)\big)
    &=\mathcal A(x)\cup\{v\},\\[-2pt]
    \mathcal F\big(x\oplus(v,m)\big)
    &=\big(\mathcal F(x)\setminus\{v\}\big)
    \cup\mathrm{ch}_m(v).
    \end{aligned}
    \label{eq:oct_transition}
\end{equation}
Starting from the root-only state, generation
ends when $\mathcal S(x)=\varnothing$, and
the remaining frontier gives the vertices.
We write $x\preceq T$ if all revealed patterns
in $x$ agree with $T$.

\noindent\textbf{Conditional Expansion Process.}
Generating a target octree $T$ takes
$R_T=|\mathcal I(T)|$ expansions, each revealing its
pattern $m_v(T)$, under topological orderings
$\mathrm{Topo}(T)$ of $\mathcal I(T)$ (each node following its parent).
By default, we use \emph{uniform topological
expansion}, $\pi\sim\mathrm{Uniform}(\mathrm{Topo}(T))$,
favoring no valid order. Given $\mathcal A(x)$, the
remaining expansions form a forest rooted at
$\mathcal S(x)$, and by the hook-length
formula~\citep{knuth1998art}, a uniformly random
topological ordering of this forest starts with root
$v$ with probability $n_v(T)/(R_T-|\mathcal A(x)|)$,
where the \emph{subtree size} $n_v(T)$ counts internal
nodes of $T$ in the subtree of $v$ (including $v$).
Proposition~\ref{prop:growth} thus gives
\begin{equation}
    r_t\big(x,x\oplus(v,m)\mid T\big)
    =\alpha_t n_v(T)\,
    \mathbb I[m=m_v(T)],
    \label{eq:oct_cond}
\end{equation}
for $x\preceq T$, $v\in\mathcal S(x)$,
and zero otherwise: each frontier node expands at rate
$n_v(T)$, prioritizing detail-rich subtrees across depths
without level-by-level barriers (Fig.~\ref{fig:octree_growth}),
with canonical rates summing to $R_T-|\mathcal A(x)|$
(Appendix~\ref{app:oct_rate}).

\noindent\textbf{Transition Rate Parameterization.}
By Eq.~\eqref{eq:pre_marginal}, the marginal rate of
expanding $v\in\mathcal S(x)$ with pattern $m$ under
uniform topological expansion is
$\alpha_t\,\mathbb E
[n_v(T)\,\mathbb I[m=m_v(T)]\mid X_t=x,c]$.
We parameterize it by
\begin{equation}
    r_t^\theta\big(x,x\oplus(v,m)\big)
    =\alpha_t\lambda_v^\theta Q_v^\theta(m),
    \label{eq:oct_param}
\end{equation}
for $v\in\mathcal S(x)$ and zero otherwise, where the
intensity $\lambda_v^\theta>0$ is the
total canonical rate of expanding $v$ and
$Q_v^\theta$ is a categorical distribution over the
255 patterns, both predicted per expandable node
(Sec.~\ref{sec:arch}). Under competing exponential
clocks, $\lambda_v^\theta$ governs expansion timing and
$Q_v^\theta$ selects $m$.

\noindent\textbf{Training and Sampling.}
For $x\preceq T$, matching the canonical rates of
Eqs.~\eqref{eq:oct_cond} and~\eqref{eq:oct_param}
under $D_\phi$ gives
\begin{equation}
    \ell_{\mathrm{oct}}\simeq\sum_{v\in\mathcal S(x)}
    \big[\lambda_v^\theta-n_v\log\lambda_v^\theta
    -n_v\log Q_v^\theta(m_v)\big],
    \label{eq:oct_loss}
\end{equation}
where $\simeq$ drops terms independent of $\theta$,
$n_v=n_v(T)$, and $m_v=m_v(T)$. Because a node fires at
rate $n_v$, the optimal factors recover the expected
subtree size and weighted pattern posterior
(Appendix~\ref{app:oct_factor_loss}):
\begin{equation}
    \begin{aligned}
    \lambda_v^*&=\mathbb E[n_v\mid X_t=x,c],\\[-2pt]
    Q_v^*(m)&=\frac{\mathbb E[n_v\mathbb I[m=m_v]\mid X_t=x,c]}
    {\lambda_v^*}.
    \end{aligned}
    \label{eq:oct_opt}
\end{equation}
We minimize
$\mathbb E_{(T,c),t,\,x\sim p_t(\cdot\mid T)}[\ell_{\mathrm{oct}}]$
via count-first sampling, sorting cumulative keys
$\sigma_v=\sigma_{\mathrm{pa}(v)}+\varepsilon_v$ ($\varepsilon_v\sim\mathrm{Exp}(n_v)$, parent $\mathrm{pa}(v)$) for $\pi$ and skipping fully expanded states
(Appendix~\ref{app:oct_count_first}).

At inference, starting from the root-only state, we
apply adaptive frozen-rate $\tau$-leaping in hazard time
$\tau_t$: in a step $h$, each $v\in\mathcal S(x)$ expands
independently with probability $1-e^{-h\lambda_v^\theta}$
and draws its pattern from $Q_v^\theta$, while newly
created children become eligible in the next step.
With $\Lambda=\sum_{v\in\mathcal S(x)}\lambda_v^\theta$,
$h$ is chosen on a physical-time grid to bound expected
expansions per step by $M$ and the missed-expansion ratio
of new children by $\epsilon$
(Appendix~\ref{app:oct_sampler}):
\begin{equation}
    \frac{h}{2\Lambda}\sum_{v\in\mathcal S(x)}
    \lambda_v^\theta\big(\lambda_v^\theta-1\big)_+
    \le\epsilon,
    \label{eq:oct_missed}
\end{equation}
where $(z)_+=\max(z,0)$.

\noindent\textbf{Breadth-First Variant.}
As a coarse-to-fine ablation, restricting $\pi$ to
level-by-level orders expands only minimal-depth
frontier nodes $\mathcal S_{\min}(x)\subseteq\mathcal S(x)$
at canonical rate $(R_T-|\mathcal A(x)|)/|\mathcal S_{\min}(x)|$
(Appendix~\ref{app:oct_bfs}).

\subsection{Edge Flow}
\label{sec:edge_flow}

\noindent\textbf{Connectivity Generation via Edge Insertion.}
Given the vertices $V=\{p_i\}_{i=1}^N$, Edge Flow
generates the mesh connectivity
$E_1\subseteq\mathcal P=\{\{i,j\}:1\le i<j\le N\}$
with a CTMC on graphs over $V$ that only inserts edges
(Fig.~\ref{fig:pipeline}, top right), conditioned on
$(V,c)$.
Starting from $E_0=\varnothing$, a state
$E\subseteq\mathcal P$ stores only inserted edges as a
sparse graph, and a jump $E\to E\cup\{e\}$ inserts a
candidate pair $e\notin E$.

\noindent\textbf{Conditional Insertion Process.}
Unlike dense graph diffusion over all $\binom{N}{2}$
pairs~\citep{vignac2023digress,qin2025defog}, only the
$|E_1|=O(N)$ mesh edges undergo transitions. Since
insertions have no precedence constraints, we draw $\pi$
uniformly over all orders of $E_1$; each missing target
edge comes next with probability $1/|E_1\setminus E|$,
and Proposition~\ref{prop:growth} gives
\begin{equation}
    r_t\big(E,E\cup\{e\}\mid E_1\big)
    =\alpha_t\,\mathbb I[e\in E_1\setminus E].
    \label{eq:edge_cond}
\end{equation}
Equivalently, target edges have independent insertion
times with distribution function $\kappa_t$, as in
masked diffusion~\citep{shi2024simplified} with its
uniformly random generation
order~\citep{hoogeboom2022autoregressive}, so $E_t$
keeps each target edge independently with probability
$\kappa_t$, and a candidate $e\notin E$ has marginal rate
$\alpha_t\Pr(e\in E_1\mid E_t=E,V,c)$ by Eq.~\eqref{eq:pre_marginal}.

\noindent\textbf{Vertex-Centered Rate Parameterization.}
Each vertex connects to only a few of its $N-1$
candidates, so most pair rates vanish whereas missing
per-vertex degrees remain small counts. We therefore
let each vertex propose edges, factorizing rates into an
intensity and a categorical choice as in Octree Flow:
$\{i,j\}$ is inserted at rate $\alpha_t w_{ij}^\theta$ with
\begin{equation}
    w_{ij}^\theta
    =\lambda_i^\theta Q_i^\theta(j)
    +\lambda_j^\theta Q_j^\theta(i),
    \label{eq:edge_param}
\end{equation}
where $\lambda_i^\theta>0$ is the canonical rate at
which vertex $i$ initiates an insertion and
$Q_i^\theta$ is a pointer over its candidate partners
$\mathcal N_i(E)=\{j\ne i:\{i,j\}\notin E\}$, a softmax
over $q_i^\top k_j/\sqrt d$ with queries and
keys $q_i,k_j\in\mathbb R^d$ (Sec.~\ref{sec:arch}).
The two terms sum proposals from either endpoint. As
each pointer is normalized,
$\sum_{\{i,j\}\notin E}w_{ij}^\theta
=\sum_{i:\mathcal N_i(E)\ne\varnothing}\lambda_i^\theta$:
intensities set total insertion counts while pointers
rank candidates. Any symmetric rates with positive row
sums take this form.

\noindent\textbf{Training and Sampling.}
We match each canonical directed term of
Eq.~\eqref{eq:edge_param} to half the target rate: with
$y_{ij}=\mathbb I[\{i,j\}\in E_1]$, matching
$\lambda_i^\theta Q_i^\theta(j)$ ($j\in\mathcal N_i(E)$)
to $\tfrac12 y_{ij}$ under Eq.~\eqref{eq:pre_kl} gives
\begin{equation}
    \begin{aligned}
    \ell_{\mathrm{edge}}\simeq{}&\sum_{i:\mathcal N_i(E)\ne\varnothing}
    \lambda_i^\theta\\[-2pt]
    &-\tfrac12\!\!
    \sum_{\{i,j\}\in E_1\setminus E}\!\!
    \log\big[\lambda_i^\theta Q_i^\theta(j)\,
    \lambda_j^\theta Q_j^\theta(i)\big].
    \end{aligned}
    \label{eq:edge_loss}
\end{equation}
This turns imbalanced $O(N^2)$ pair regression into
per-vertex missing-degree estimation and local softmax
competition over each endpoint's candidates, minimized in
expectation over $(V,E_1,c)$, $t$, and
$E\sim p_t(\cdot\mid E_1)$. With pair loss
$\ell_{\mathrm{pair}}=\sum_{\{i,j\}\notin E}
D_\phi(y_{ij},w_{ij}^\theta)$ and endpoint split
$s_{ij}=\lambda_i^\theta Q_i^\theta(j)/w_{ij}^\theta$,
\begin{equation}
    \ell_{\mathrm{edge}}
    \simeq\ell_{\mathrm{pair}}
    +\!\sum_{\{i,j\}\in E_1\setminus E}\!\!
    \mathrm{KL}\big((\tfrac12,\tfrac12)\,\|\,(s_{ij},1-s_{ij})\big)
    \label{eq:edge_balance_main}
\end{equation}
(Appendix~\ref{app:edge_loss}), so the optimum splits
the marginal canonical pair rate equally across
endpoints, with $\lambda_i^*=\tfrac12\mathbb E[\delta_i\mid E_t=E,V,c]$
and $Q_i^*(j)=\Pr(\{i,j\}\in E_1\mid E_t=E,V,c)/(2\lambda_i^*)$ ($\lambda_i^*>0$)
for missing degree $\delta_i=\sum_{j\in\mathcal N_i(E)}y_{ij}$.

At inference, starting from $E_0=\varnothing$ on a
uniform grid $0=t_0<\cdots<t_S=1$, a missing target
edge arrives in $[t_k,t_{k+1}]$ with probability
$\gamma_k=(\kappa_{t_{k+1}}-\kappa_{t_k})/(1-\kappa_{t_k})$.
Estimating $w=\Pr(\{i,j\}\in E_1\mid E_t=E,V,c)$ by
$\hat w_{ij}=\operatorname{clip}(w_{ij}^\theta,0,1)$ at
$t_k$, we insert each candidate pair independently with
probability
\begin{equation}
    \begin{aligned}
    P_k(i,j)&=\gamma_k\,\hat w_{ij}^\beta,\\[-2pt]
    \hat w_{ij}^\beta&=\sigma\!\left(
    \frac{\operatorname{logit}\hat w_{ij}-\log(1-\kappa_{t_k})}{\beta}
    +\log(1-\kappa_{t_k})\right),
    \end{aligned}
    \label{eq:edge_sample}
\end{equation}
for $\hat w_{ij}\in(0,1)$ ($\hat w_{ij}^\beta=\hat w_{ij}$ at $\{0,1\}$),
sigmoid $\sigma$, and temperature $\beta>0$.
At $\beta=1$, $\hat w_{ij}^1=\hat w_{ij}$ gives the exact
ancestral pair marginal ($\gamma_{S-1}=1$ decides
remaining pairs); for $\beta\ne 1$, subtracting
$\log(1-\kappa_t)$ in
$\operatorname{logit}w=\operatorname{logit}\tilde w+\log(1-\kappa_t)$
sharpens the other-pair belief $\tilde w$ without shifting timing
(Appendix~\ref{app:edge_sampling}). Faces are extracted as 3-cliques.

\subsection{Network Architecture}
\label{sec:arch}

Both stages use Transformers over 3D spatial tokens
(Fig.~\ref{fig:pipeline}, bottom). Each block applies
self-attention with 3D rotary positional embeddings,
cross-attention to condition tokens $c$, and
time-modulated normalization. Condition $c$ comes
from a jointly trained VecSet encoder~\citep{zhang20233dshape2vecset}
for point clouds or frozen DINOv2-L~\citep{oquab2024dinov2}
for images (Appendix~\ref{app:impl}).

\noindent\textbf{Octree Flow} represents a partial tree
by its frontier $\mathcal F(x)$, embedding each node's
octant address and depth at its cell center to predict
$\lambda_v^\theta$ (via $\exp$) and 255-way $Q_v^\theta$
on $\mathcal S(x)$.

\noindent\textbf{Edge Flow} embeds each vertex's
coordinates and current degree, adds message passing
along inserted edges with relative-position features after
cross-attention, and predicts $\lambda_i^\theta$ (via $\exp$)
with queries and keys ($d=128$) for $Q_i^\theta$.

\section{Experiments}
\label{sec:experiments}

\begin{figure*}[!t]
    \centering
    \includegraphics[width=0.91\textwidth]{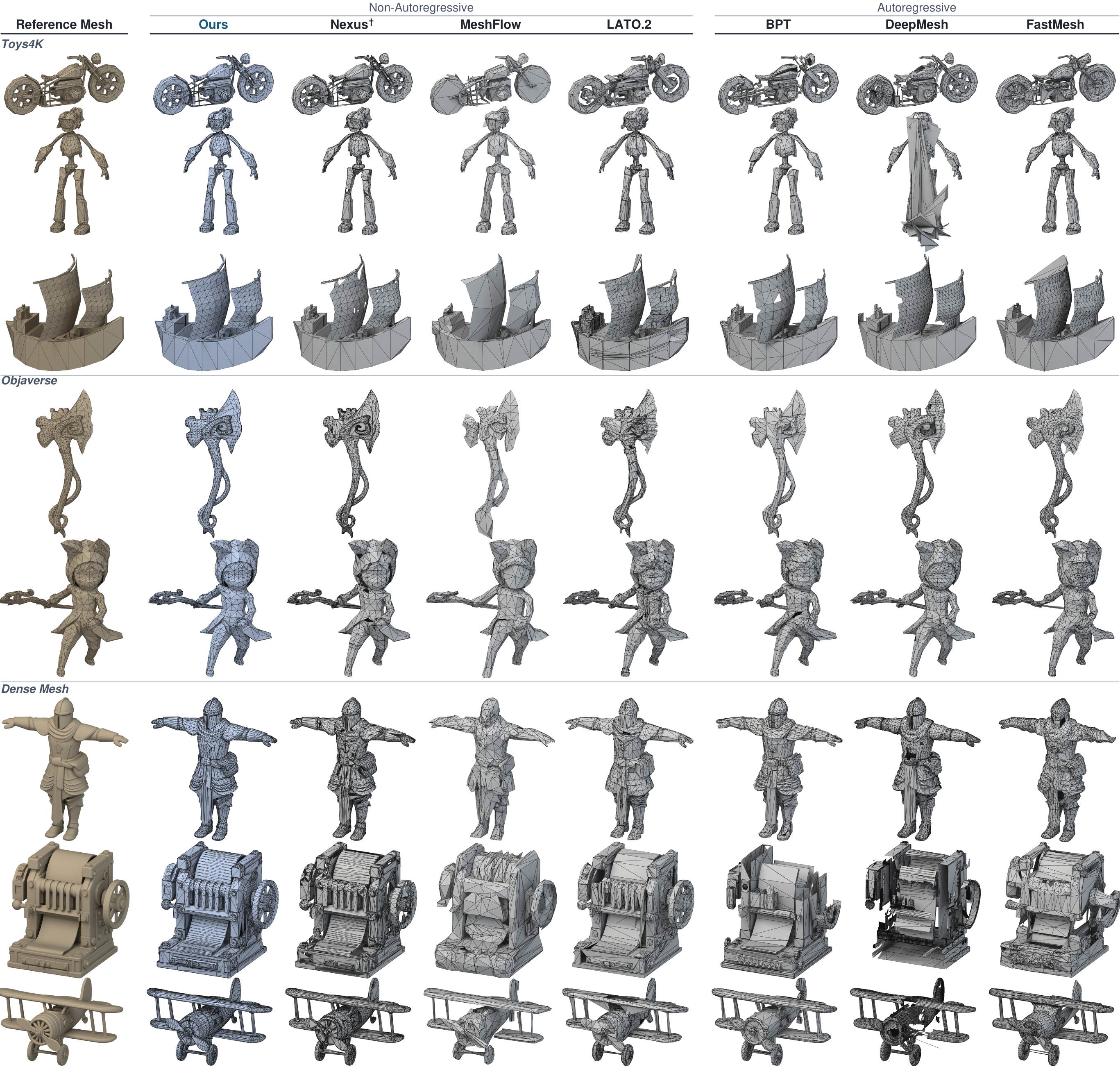}%
    \vspace{1.2mm}
    \caption{\textbf{Point-cloud-conditioned mesh generation} on Toys4K, Objaverse, and Dense Mesh~\citep{xiang2025trellis2,lai2026lattice,jia2025ultrashape} ($^\dagger$controlled reimplementation).}
    \label{fig:qualitative_comparison}
\end{figure*}

\subsection{Implementation Details}
\label{sec:exp_setup}

\paragraph{Models and Training.}
We train \methodname on $\sim\!1.17\text{M}$ filtered Objaverse-XL~\citep{deitke2023objaversexl} artistic meshes ($<8\text{K}$ vertices) in $[-1,1]^3$ quantized onto a $1024^3$ octree ($L=10$).
Octree Flow ($\mbox{$\sim\!370\text{M}$}$/$\mbox{$\sim\!670\text{M}$}$ parameters for point clouds/images) trains for $500\text{K}$ steps under $\kappa_t=t^3$ (allocating resolution to early branching), and Edge Flow ($\mbox{$\sim\!700\text{M}$}$) for $200\text{K}$ steps under $\kappa_t=t$.
Inference uses adaptive $\tau$-leaping ($50$-step grid, $\epsilon=0.05$, $M=128$, pattern temperature $0.1$; classifier-free guidance (CFG) weight $0$ on Toys4K/Objaverse, $1.0$ on Dense Mesh, $2.0$ for images) for Octree Flow and $S=50$ uniform steps ($\beta=0.3$, no CFG) for Edge Flow.
As Nexus~\citep{wang2026nexus} lacks public code, we reimplement both stages (\nexus) under matched data, conditioner, backbones, batch size, and step budgets, with its topology VAE (Appendix~\ref{app:impl}).

\paragraph{Benchmarks and Metrics.}
We evaluate on held-out \textbf{Toys4K} ($1{,}068$ shapes), \textbf{Objaverse} ($500$ shapes), and \textbf{Dense Mesh} ($222$ generated dense meshes~\citep{xiang2025trellis2,lai2026lattice,jia2025ultrashape}); factorized analyses and ablations use $40\text{K}$ points unless noted.
We report symmetric $L_1$ Chamfer distance (\textbf{CD}, $\times 10^3$), Hausdorff distance (\textbf{HD}, $\times 10^2$), and normal consistency (\textbf{NC}) over $10^5$ surface samples, plus undirected \textbf{Edge F1} (\%) on ground-truth vertices (Appendix~\ref{app:benchmarks}).

\subsection{Mesh Generation}
\label{sec:exp_main}

\begin{table}[t]
    \centering
    \caption{\textbf{Point-cloud-conditioned mesh generation} (16K/40K points; $^\dagger$controlled reimplementation; $^\ddagger$image + sparse voxels).}
    \label{tab:main}
    \setlength{\tabcolsep}{2.2pt}
    \resizebox{\columnwidth}{!}{%
    \begin{tabular}{lccccccccc}
        \toprule
        & \multicolumn{3}{c}{Toys4K} & \multicolumn{3}{c}{Objaverse} & \multicolumn{3}{c}{Dense Mesh} \\
        \cmidrule(lr){2-4}\cmidrule(lr){5-7}\cmidrule(lr){8-10}
        Method & CD$\downarrow$ & HD$\downarrow$ & NC$\uparrow$
        & CD$\downarrow$ & HD$\downarrow$ & NC$\uparrow$
        & CD$\downarrow$ & HD$\downarrow$ & NC$\uparrow$ \\
        \midrule
        \multicolumn{10}{l}{\textit{Autoregressive}} \\
        EdgeRunner~\citep{tang2025edgerunner} & 15.26 & 17.12 & 0.905 & 18.92 & 20.92 & 0.894 & 23.65 & 31.09 & 0.794 \\
        TreeMeshGPT~\citep{lionar2025treemeshgpt} & 10.51 & 13.91 & 0.914 & 12.04 & 17.51 & 0.904 & 42.82 & 39.97 & 0.766 \\
        BPT~\citep{weng2025bpt} & 6.36 & 8.26 & 0.931 & 6.67 & 11.04 & 0.925 & 57.23 & 48.80 & 0.801 \\
        DeepMesh~\citep{zhao2025deepmesh} & 8.02 & 11.47 & 0.939 & 8.28 & 14.55 & 0.937 & 45.74 & 42.96 & 0.810 \\
        FastMesh~\citep{kim2025fastmesh} & 7.86 & 9.08 & 0.899 & 12.83 & 14.54 & 0.883 & 45.70 & 26.03 & 0.715 \\
        MeshRipple~\citep{lin2026meshripple} & 7.26 & 10.20 & 0.919 & 8.74 & 14.62 & 0.912 & 35.91 & 38.68 & 0.762 \\
        \midrule
        \multicolumn{10}{l}{\textit{Non-autoregressive}} \\
        MeshFlow~\citep{li2026meshflow} & 9.75 & 6.45 & 0.893 & 13.45 & 8.35 & 0.878 & 12.04 & \underline{8.56} & 0.803 \\
        LATO.2$^\ddagger$~\citep{long2026lato2} & 5.10 & 6.73 & 0.897 & 6.37 & 8.12 & 0.885 & 6.33 & 9.70 & 0.826 \\
        \nexus~\citep{wang2026nexus} (16K) & 2.67 & 8.47 & 0.961 & 3.97 & 11.42 & 0.955 & 4.41 & 12.08 & 0.887 \\
        \nexus~\citep{wang2026nexus} (40K) & 2.24 & 7.32 & 0.967 & 2.99 & 9.25 & 0.960 & 4.05 & 11.45 & 0.892 \\
        \methodname (16K) & \underline{1.56} & \underline{4.06} & \underline{0.980} & \underline{1.95} & \underline{5.75} & \underline{0.976} & \underline{3.02} & 10.20 & \underline{0.937} \\
        \methodname (40K) & \textbf{1.46} & \textbf{3.88} & \textbf{0.981} & \textbf{1.52} & \textbf{4.68} & \textbf{0.978} & \textbf{2.22} & \textbf{8.54} & \textbf{0.940} \\
        \bottomrule
    \end{tabular}}
    \vspace{-1mm}
\end{table}

\begin{figure*}[!t]
    \centering
    \includegraphics[width=0.995\textwidth]{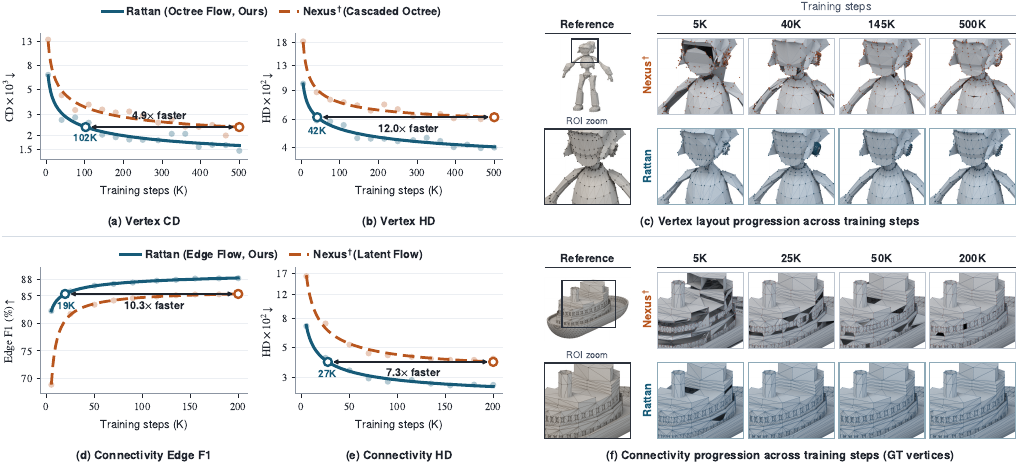}%
    \vspace{0.5mm}
    \caption{\textbf{Controlled two-stage training convergence on Toys4K} ($^\dagger$controlled reimplementation):
    vertex generation wired by frozen Edge Flow (\textbf{a--c}) and connectivity generation on ground-truth vertices (\textbf{d--f}).}
    \label{fig:convergence}
\end{figure*}

\paragraph{Quantitative Comparison.}
Tab.~\ref{tab:main} compares \methodname with autoregressive~\citep{tang2025edgerunner,lionar2025treemeshgpt,weng2025bpt,zhao2025deepmesh,kim2025fastmesh,lin2026meshripple} and non-autoregressive~\citep{li2026meshflow,long2026lato2,wang2026nexus} baselines on Toys4K, Objaverse, and Dense Mesh.
Existing methods reveal a clear paradigm trade-off: compressive tokenization (BPT~\citep{weng2025bpt}) and RL post-training (DeepMesh~\citep{zhao2025deepmesh}) improve CD and NC among sequential decoders on bounded benchmarks, yet token budgets and autoregressive drift keep worst-case HD high ($>8.2$ on Toys4K; $>11.0$ on Objaverse) and truncate complex Dense Mesh shapes (CD $>23.6$, HD $>26.0$), while one-shot wiring in FastMesh~\citep{kim2025fastmesh} adds redundant faces that hurt NC ($0.899/0.883/0.715$).
Conversely, parallel latent models (MeshFlow~\citep{li2026meshflow}, LATO.2~\citep{long2026lato2}) curb global HD drift, but lossy latent decoding oversimplifies geometry or yields jagged facets, inflating CD ($5.10\text{--}13.45$) and depressing NC ($0.803\text{--}0.897$); \nexus~\citep{wang2026nexus} improves CD and NC via octree cascades and spacetime latents, yet retains elevated HD ($7.32/9.25/11.45$) from relaxed-occupancy outliers and decoding defects.
Growing octrees and edges natively resolves both bottlenecks: \methodname achieves the best score on every metric across all benchmarks, cutting CD by ${4.4\times}\text{--}{10.7\times}$ over the strongest autoregressive baselines, lowering CD by $34.8\%/49.2\%/45.2\%$ and HD by $47.0\%/49.4\%/25.4\%$ against $40\text{K}$-point \nexus, and surpassing $40\text{K}$-point \nexus even with $16\text{K}$ points.

\begin{figure}[t]
    \centering
    \includegraphics[width=0.975\columnwidth]{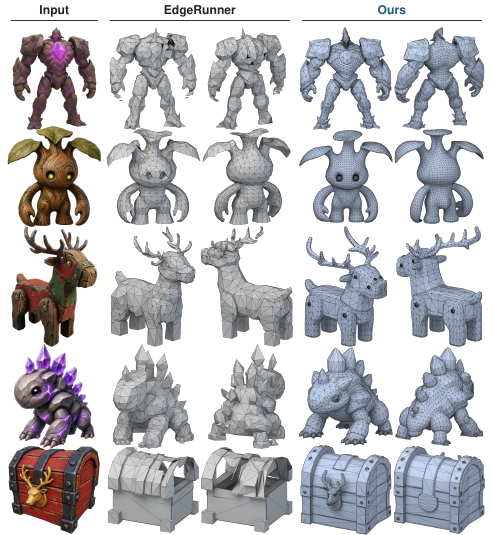}%
    \caption{\textbf{Single-image-to-mesh generation} vs.\ EdgeRunner~\citep{tang2025edgerunner}.}
    \label{fig:image_to_mesh}
\end{figure}

\paragraph{Qualitative Comparison.}
Visual comparisons in Fig.~\ref{fig:qualitative_comparison} link these metrics to distinct structural artifacts.
Under sequence budgets and autoregressive drift, sequential decoders either under-tessellate carvings and drop delicate parts such as the sailboat rigging and biplane struts (BPT~\citep{weng2025bpt}) or produce runaway stray polygons and unclosed cavities on complex shapes like the robot and machinery (DeepMesh~\citep{zhao2025deepmesh}), whereas one-shot wiring in FastMesh~\citep{kim2025fastmesh} stacks redundant faces into cluttered wireframes.
Non-autoregressive baselines avoid sequential drift, yet continuous relaxations and latent autoencoding still degrade mesh structure: joint latents in MeshFlow~\citep{li2026meshflow} oversimplify fine geometry, while two-stage methods yield jagged, overlapping facets in LATO.2~\citep{long2026lato2} or uneven vertex clustering from relaxed cascades and local punctures from latent wiring in \nexus~\citep{wang2026nexus} (e.g., on the axe and sailboat).
In contrast, \methodname recovers complete surfaces with clean, feature-aligned triangulations, crisp creases, and intact thin structures, distilling over-tessellated ($\sim\!500\text{K}$-face) generated isosurfaces~\citep{xiang2025trellis2,lai2026lattice,jia2025ultrashape} (bottom) into compact, crease-preserving artistic meshes.

\paragraph{Image-Conditioned Generation.}
On single-image inputs under monocular ambiguity (Fig.~\ref{fig:image_to_mesh}), EdgeRunner~\citep{tang2025edgerunner} produces coarse, fragmented meshes with shoulder holes and a shattered chest lid, whereas \methodname recovers complete front-and-back topology, smooth curvature, and sharp details such as branched antlers, dorsal crystals, and the stag emblem.

\subsection{Factorized and Convergence Analysis}
\label{sec:exp_factorized}

\begin{table}[t]
    \centering
    \caption{\textbf{Connectivity generation from ground-truth vertices.}}
    \label{tab:gt_vertices}
    \setlength{\tabcolsep}{3pt}
    \resizebox{\columnwidth}{!}{%
    \begin{tabular}{lcccccccc}
        \toprule
        & \multicolumn{4}{c}{Toys4K} & \multicolumn{4}{c}{Objaverse} \\
        \cmidrule(lr){2-5}\cmidrule(lr){6-9}
        Method & CD$\downarrow$ & HD$\downarrow$ & NC$\uparrow$ & F1$\uparrow$
        & CD$\downarrow$ & HD$\downarrow$ & NC$\uparrow$ & F1$\uparrow$ \\
        \midrule
        LATO.2~\citep{long2026lato2} & 3.03 & 8.18 & 0.962 & 78.3 & 6.54 & 13.24 & 0.936 & 71.9 \\
        FastMesh~\citep{kim2025fastmesh} & 1.57 & 6.10 & 0.974 & 78.6 & 3.51 & 10.33 & 0.958 & 73.6 \\
        \nexus~\citep{wang2026nexus} & \underline{0.92} & \underline{4.30} & \underline{0.990} & \underline{85.3} & \underline{2.00} & \underline{6.68} & \underline{0.976} & \underline{80.4} \\
        \methodname & \textbf{0.71} & \textbf{2.83} & \textbf{0.992} & \textbf{88.3} & \textbf{0.98} & \textbf{3.61} & \textbf{0.986} & \textbf{84.5} \\
        \bottomrule
    \end{tabular}}
    \vspace{-0.7mm}
\end{table}

\begin{table}[t]
    \centering
    \caption{\textbf{Controlled stage-wise comparison on Toys4K.}}
    \label{tab:stage_swap}
    \setlength{\tabcolsep}{4.5pt}
    \resizebox{\columnwidth}{!}{%
    \begin{tabular}{llccc}
        \toprule
        Vertex Stage & Connectivity Stage & CD$\downarrow$ & HD$\downarrow$ & NC$\uparrow$ \\
        \midrule
        \nexus (Cascaded) & \nexus (Latent Flow) & 2.24 & 7.32 & 0.967 \\
        \nexus (Cascaded) & \methodname (Edge Flow) & 2.37 & 6.45 & 0.974 \\
        \methodname (Octree Flow) & \nexus (Latent Flow) & \underline{1.63} & \underline{5.25} & \underline{0.978} \\
        \methodname (Octree Flow) & \methodname (Edge Flow) & \textbf{1.46} & \textbf{3.88} & \textbf{0.981} \\
        \bottomrule
    \end{tabular}}
    \vspace{-1mm}
\end{table}

\paragraph{Connectivity Generation.}
To isolate Edge Flow from vertex errors, Tab.~\ref{tab:gt_vertices} evaluates connectivity on ground-truth vertices.
FastMesh~\citep{kim2025fastmesh} predicts edges in one pass and introduces stray faces that hurt HD, while LATO.2~\citep{long2026lato2} and \nexus~\citep{wang2026nexus} route discrete adjacency through continuous latents.
Edge Flow surpasses \nexus by ${+3.0}/{+4.1}$ Edge F1 points ($88.3\%/84.5\%$) and lowers HD by $34.2\%/46.0\%$ on Toys4K and Objaverse via iterative graph edge insertion.

\paragraph{Stage-Wise Contribution.}
Tab.~\ref{tab:stage_swap} evaluates all four vertex--connectivity pairings of \nexus and \methodname on Toys4K.
Under either connectivity model, replacing \nexus's cascaded octree generator with Octree Flow improves all metrics, cutting CD from $2.24$ to $1.63$ with Latent Flow and from $2.37$ to $1.46$ with Edge Flow.
Conversely, Edge Flow improves HD and NC over Latent Flow on both vertex sets and improves all metrics on Octree Flow's vertices (HD $-26.1\%$, $5.25\to 3.88$).

\paragraph{Training Convergence.}
Fig.~\ref{fig:convergence} evaluates 15 vertex ($5\text{K}$--$500\text{K}$ steps) and 10 connectivity ($5\text{K}$--$200\text{K}$ steps) checkpoints on Toys4K.
For vertices (\textbf{a--c}), Octree Flow matches $500\text{K}$-step \nexus in $102\text{K}$ steps on CD ($4.9\times$ faster) and $42\text{K}$ on HD ($12.0\times$ faster), rapidly forming compact, feature-aligned layouts whereas cascaded relaxation scatters off-surface vertices early and retains uneven clusters at $500\text{K}$.
For connectivity (\textbf{d--f}), Edge Flow, trained from scratch without a topology VAE, reaches $82.3\%$ Edge F1 by $5\text{K}$ steps (vs.\ $68.9\%$ for Latent Flow) and matches $200\text{K}$-step \nexus in $19\text{K}$ steps on F1 ($10.3\times$ faster) and $27\text{K}$ on HD ($7.3\times$ faster), recovering coherent wiring already at $5\text{K}$ steps.

\subsection{Ablation Studies}
\label{sec:exp_ablation}

\begin{table}[t]
    \centering
    \caption{\textbf{Octree generation ablation} (Edge Flow; H20 runtime).}
    \label{tab:ablation_octree}
    \setlength{\tabcolsep}{2.5pt}
    \resizebox{\columnwidth}{!}{%
    \begin{tabular}{lcccccccc}
        \toprule
        & \multicolumn{3}{c}{Toys4K} & \multicolumn{3}{c}{Objaverse} & \multicolumn{2}{c}{Efficiency} \\
        \cmidrule(lr){2-4}\cmidrule(lr){5-7}\cmidrule(lr){8-9}
        Vertex Generator & CD$\downarrow$ & HD$\downarrow$ & NC$\uparrow$
        & CD$\downarrow$ & HD$\downarrow$ & NC$\uparrow$ & NFE$\downarrow$ & Time (s)$\downarrow$ \\
        \midrule
        \nexus (Cascaded) & 2.37 & 6.45 & 0.974 & 2.64 & 6.64 & 0.971 & 200.0 & 7.75 \\
        Breadth-first (App.~\ref{app:oct_bfs}) & \underline{1.75} & \underline{4.20} & \underline{0.979} & \underline{1.78} & \underline{5.58} & \underline{0.976} & \textbf{89.2} & \textbf{3.49} \\
        Uniform (Eq.~\eqref{eq:oct_cond}) & \textbf{1.46} & \textbf{3.88} & \textbf{0.981} & \textbf{1.52} & \textbf{4.68} & \textbf{0.978} & \underline{136.6} & \underline{5.15} \\
        \bottomrule
    \end{tabular}}
\end{table}

\paragraph{Octree Expansion Process.}
Tab.~\ref{tab:ablation_octree} isolates discrete transitions and cross-depth expansion against \nexus's cascaded continuous octree flow under the same frozen Edge Flow.
At matched level-wise order, discrete breadth-first growth (Appendix~\ref{app:oct_bfs}) surpasses continuous relaxation on all metrics while running $2.2\times$ faster ($3.49\text{s}$ vs.\ $7.75\text{s}$; $89.2$ vs.\ $200$ NFEs).
Uniform topological expansion (Eq.~\eqref{eq:oct_cond}) lifts the depth barrier to prioritize detail-rich subtrees (larger $n_v(T)$), cutting CD by $16.6\%/14.6\%$ ($1.46/1.52$) and HD by $7.6\%/16.1\%$ at $1.5\times$ speedup ($5.15\text{s}$, $136.6$ NFEs).

\begin{table}[t]
    \centering
    \caption{\textbf{Edge Flow ablation} (image, GT vertices, Toys4K).}
    \label{tab:ablation_edge}
    \setlength{\tabcolsep}{4pt}
    \resizebox{\columnwidth}{!}{%
    \begin{tabular}{llcccc}
        \toprule
        Rate Parameterization & Temp.\ $\beta$ & CD$\downarrow$ & HD$\downarrow$ & NC$\uparrow$ & F1$\uparrow$ \\
        \midrule
        Direct pairwise rate & $0.3$ & 3.95 & 10.11 & 0.969 & 82.9 \\
        Direct pairwise rate & $0.5$ & 3.54 & 9.49 & 0.971 & 82.7 \\
        \midrule
        Vertex-centered (Eq.~\eqref{eq:edge_param}) & $1.0$ & 3.49 & 11.12 & 0.964 & 80.4 \\
        Vertex-centered (Eq.~\eqref{eq:edge_param}) & $0.7$ & \underline{3.04} & 9.51 & \underline{0.972} & 82.3 \\
        Vertex-centered (Eq.~\eqref{eq:edge_param}) & $0.5$ & \textbf{2.98} & \textbf{8.90} & \textbf{0.974} & 83.0 \\
        Vertex-centered (Eq.~\eqref{eq:edge_param}) & $0.3$ & 3.44 & \underline{9.07} & \underline{0.972} & \underline{83.2} \\
        Vertex-centered (Eq.~\eqref{eq:edge_param}) & $0.1$ & 3.59 & 9.39 & 0.971 & \textbf{83.7} \\
        \bottomrule
    \end{tabular}}
\end{table}

\paragraph{Edge Rate Parameterization.}
Tab.~\ref{tab:ablation_edge} compares vertex-centered rates (Eq.~\eqref{eq:edge_param}) with a matched baseline predicting symmetric pair rates directly via an exponential link on dot-product logits.
At matched $\beta$, decoupling vertex intensities $\lambda_i^\theta$ from candidate pointers $Q_i^\theta$ improves every metric, cutting CD by $12.9\%$ at $\beta=0.3$ ($3.44$ vs.\ $3.95$) and $15.8\%$ at $\beta=0.5$ ($2.98$ vs.\ $3.54$), and HD to $9.07/8.90$.

\paragraph{Sampling Temperature.}
Tab.~\ref{tab:ablation_edge} (bottom) sweeps sharpening temperature $\beta$ (Appendix~\ref{app:edge_sampling}), concentrating pair beliefs without shifting arrival rates.
Lowering $\beta$ ($1.0\to 0.1$) monotonically raises F1 ($80.4\%\to 83.7\%$) by pruning uncertain edges: $\beta=1.0$ admits false positives ($11.12$ HD), while $\beta=0.1$ omits faces ($3.59$ CD, $9.39$ HD).
While $\beta=0.5$ lowers surface error on GT vertices, we default to $\beta=0.3$ ($83.2\%$ F1) to suppress ambiguous edges and non-manifold 3-cliques when wiring predicted vertices.

\section{Conclusion}
\label{sec:conclusion}

We introduced \methodname, a native mesh generator built on \emph{Growth Flows}, CTMCs for discrete growth with closed-form conditional rates: \emph{Octree Flow} for variable-cardinality vertices and vertex-centered \emph{Edge Flow} for connectivity enable parallel synthesis directly on octrees and graphs, improving quality, convergence, and dense-mesh conversion.

\paragraph{Limitations and Future Work.}
As a pure insertion process, Edge Flow cannot retract early false-positive edges, and taking 3-cliques of $E$ as faces leaves orientations unspecified and may include non-facial 3-cycles; both can be addressed on the generated graph by a lightweight learned network for face filtering and orientation~\citep{shen2024spacemesh,wang2026nexus}, while reversible deletions and quad-dominant meshes remain future work.

\clearpage
{
    \small
    \bibliographystyle{ieeenat_fullname}
    \bibliography{main}
}

\clearpage
\twocolumn[{%
  \centering
  \Large\textbf{Appendix}\par
  \vspace{1.0em}
}]
\appendix

\section{Growth Flows}

\subsection{CTMC Generators}
\label{app:ctmc}
A CTMC on a discrete state space $\mathcal X$ is
specified by a generator $r_t(x,y)$, whose
off-diagonal entries are nonnegative and whose
diagonal satisfies
$r_t(x,x)=-\sum_{y\ne x}r_t(x,y)$.
Its marginals $p_t$ obey the Kolmogorov forward
equation
\begin{equation}
    \partial_t p_t(x)
    =\sum_{y\in\mathcal X}p_t(y)r_t(y,x).
    \label{eq:pre_kfe}
\end{equation}

\subsection{Proof of Proposition~\ref{prop:growth}}
\label{app:growth}
Let $U$ be the set of the $R$ events of a target $x_1$.
Marks are determined by $x_1$ and revealed with their
events. For an order $\pi$ of $U$, the first $k$ events
of $\pi$ form the revealed set $A$.

We work in hazard time $\tau=\tau_t=-\log(1-\kappa_t)$,
with $d\tau_t/dt=\alpha_t$. In this time coordinate,
the i.i.d.\ timestamps are unit-rate exponential. Their
count $K_t$ below $\tau_t$ jumps $k\to k+1$ at rate
$R-k$, independently of $\pi$, which gives
\begin{equation}
 K_t\sim\mathrm{Binomial}\big(R,1-e^{-\tau_t}\big)
 =\mathrm{Binomial}(R,\kappa_t).
 \label{eq:app_growth_count}
\end{equation}
When the count jumps from $k$ to $k+1$, the revealed
event is $\pi_{k+1}$. Note that the trajectory of $X$ up
to time $t$ reveals both the ordered prefix
$(\pi_1,\ldots,\pi_k)$ and its jump times. If the law of
$\pi_{k+1}$ given $\pi_1,\ldots,\pi_k$ depends only on the unordered set
$A=\{\pi_1,\ldots,\pi_k\}$ at every $k<R$, then the law of
the remaining suffix $(\pi_{k+1},\ldots,\pi_R)$ given
the prefix also depends only on $A$, as the product of
these next-event probabilities. Moreover, the remaining
$R-k$ unit-rate exponential event times are memoryless
and independent of $\pi$ (hence of the past jump times
and prefix order). Conditioned on the trajectory up to
$t$, the future evolution therefore depends only on the
current state $X_t=A$, so $X_t$ is Markov given $x_1$.
Its canonical rate of revealing $v$ is
$(R-k)\Pr(\pi_{k+1}=v\mid A)$, which gives
Eq.~\eqref{eq:growth_rate} in physical time, and these
rates sum to $R-k$. Since $K_t$ and $\pi$ are
independent, $X_t\mid x_1$ is sampled exactly by drawing
$K\sim\mathrm{Binomial}(R,\kappa_t)$ and an independent
$\pi$, and revealing $\pi_1,\ldots,\pi_K$.
\hfill$\square$

\section{Additional Details of Octree Flow}
\label{app:octree}

Throughout, $x\preceq T$ is a partial octree
consistent with the target $T$,
$R_T(x)=R_T-|\mathcal A(x)|$ is its number of
remaining expansions, and $\mathrm{Topo}(T)$ is the
set of topological orderings of $\mathcal I(T)$, in
which every node follows its parent.

\subsection{Uniform Topological Expansion: Rates and Closed Form}
\label{app:oct_rate}

\paragraph{Remaining Forest.}
Every unexpanded internal node
$u\in\mathcal I(T)\setminus\mathcal A(x)$ has a unique
first unexpanded node $v$ on the root-to-$u$ path,
possibly $v=u$. Either $v$ is the root or its parent is
expanded, and $v$ is internal, so $v\in\mathcal S(x)$,
and $u$ lies in the target subtree of $v$. Hence the
target subtrees rooted at $\mathcal S(x)$ partition
$\mathcal I(T)\setminus\mathcal A(x)$, so
\begin{equation}
 \sum_{v\in\mathcal S(x)}n_v(T)
 =R_T-|\mathcal A(x)|.
 \label{eq:oct_conservation}
\end{equation}
The set
$\mathcal A(x)$ is closed under ancestors, and the
remaining expansions form the forest $F$ of these
subtrees, in which the hook length of every node $u$,
the size of its subtree, is $n_u(T)$.

\paragraph{Rates.}
A topological ordering of $\mathcal I(T)$ whose first
$k$ nodes form $\mathcal A(x)$ is a topological ordering
of $\mathcal A(x)$ followed by one of $F$, since no node
of $F$ precedes a node of $\mathcal A(x)$. Given
$\mathcal A(x)$, the remaining order is therefore
uniform over the topological orderings of $F$, whatever
the order of the prefix, which satisfies the condition
of Proposition~\ref{prop:growth}. By the hook-length
formula for forests~\citep{knuth1998art}, a forest $F$
has $e(F)=|F|!/\prod_{u\in F}h_F(u)$ topological
orderings, where $h_F(u)$ is the size of the subtree of
$u$ in $F$. Removing a root $v$ leaves the hook lengths
of the other nodes unchanged, so
\begin{equation}
 \Pr\big(\pi_{k+1}=v\mid\mathcal A(x)\big)
 =\frac{e(F\setminus\{v\})}{e(F)}
 =\frac{n_v(T)}{R_T(x)}
 \label{eq:app_oct_first}
\end{equation}
for $v\in\mathcal S(x)$. Proposition~\ref{prop:growth}
then gives the canonical rate
$R_T(x)\cdot n_v(T)/R_T(x)=n_v(T)$, which is
Eq.~\eqref{eq:oct_cond}.

\paragraph{Sampling Uniform Topological Orderings.}
Give every internal node $v$ with parent $\mathrm{pa}(v)$ an independent delay
$\varepsilon_v\sim\mathrm{Exp}(n_v(T))$ and set
\begin{equation}
 \sigma_v=\sigma_{\mathrm{pa}(v)}+\varepsilon_v,\qquad
 \sigma_{\mathrm{pa}(\mathrm{root})}=0.
 \label{eq:oct_keys}
\end{equation}
In key order, every node follows its parent. The nodes
whose parents have fired but which have not fired
themselves form the expandable frontier, and by
memorylessness their residual delays are independent
with rates $n_v(T)$, which sum to the number of
remaining nodes by Eq.~\eqref{eq:oct_conservation}. The
$j$-th node in key order is therefore $v$ with
probability $n_v(T)/(R_T-j+1)$, and an order
$\pi=(v_1,\ldots,v_{R_T})$ has probability
\begin{equation}
 \Pr(\pi\mid T)
 =\prod_{j=1}^{R_T}
 \frac{n_{v_j}(T)}{R_T-j+1}
 =\frac{\prod_{v\in\mathcal I(T)}n_v(T)}{R_T!},
 \label{eq:app_oct_topo}
\end{equation}
which does not depend on $\pi$. Sorting the keys thus
samples $\pi\sim\mathrm{Uniform}(\mathrm{Topo}(T))$.
The keys are also the expansion times, in hazard time,
of the chain with the rates of Eq.~\eqref{eq:oct_cond},
so thresholding them at $\tau_t$ also samples
$X_t\mid T$.

\paragraph{Closed-Form Conditional Path.}
Let $A=\mathcal A(x)$ with $|A|=k$, and let $h_x(v)$
be the number of nodes of $A$ in the subtree rooted at
$v$, including $v$. By the hook-length formula,
$e(A)=k!/\prod_{v\in A}h_x(v)$,
$e(F)=(R_T-k)!/\prod_{u\in\mathcal I(T)\setminus A}n_u(T)$,
and
$|\mathrm{Topo}(T)|=R_T!/\prod_{v\in\mathcal I(T)}n_v(T)$.
Exactly $e(A)\,e(F)$ topological orderings of
$\mathcal I(T)$ begin with $A$, and $K_t$ is independent
of $\pi$, so
\begin{equation}
 \begin{aligned}
 p_t(x\mid T)
 &=\binom{R_T}{k}\kappa_t^k(1-\kappa_t)^{R_T-k}\,
 \frac{e(A)\,e(F)}{|\mathrm{Topo}(T)|}\\[-2pt]
 &=\kappa_t^k(1-\kappa_t)^{R_T-k}
 \prod_{v\in\mathcal A(x)}
 \frac{n_v(T)}{h_x(v)}
 \end{aligned}
 \label{eq:oct_path}
\end{equation}
for $x\preceq T$ (with the empty product equal to $1$ at
$k=0$, giving $(1-\kappa_t)^{R_T}$, and $h_T(v)=n_v(T)$
at $k=R_T$, giving $\kappa_t^{R_T}$), and zero otherwise.

\subsection{Breadth-First Expansion}
\label{app:oct_bfs}

Breadth-first expansion uses the layered orders, in
which all internal nodes of a depth precede those of
the next depth, in any order within a depth. For $x\preceq T$
with $k=|\mathcal A(x)|<R_T$, let
$\ell=\min_{v\in\mathcal S(x)}\operatorname{depth}(v)$ and
$\mathcal S_{\min}(x)=\{v\in\mathcal S(x):\operatorname{depth}(v)=\ell\}$.
All nodes of depth $\ell$ are instantiated once depth $\ell-1$ is
complete, so in a consistent layered state (all depths $<\ell$
complete and no node of depth $>\ell$ expanded), $\mathcal S_{\min}(x)$
comprises all remaining internal nodes of $T$ at depth $\ell$. A layered order whose first $k$ nodes form
$\mathcal A(x)$ arranges $\mathcal A(x)$ depth by depth,
followed by a permutation of $\mathcal S_{\min}(x)$ and
of every deeper depth. Given $\mathcal A(x)$, the next
node is therefore uniform over $\mathcal S_{\min}(x)$,
whatever the order of the prefix, and
Proposition~\ref{prop:growth} gives
\begin{equation}
 r_t\big(x,x\oplus(v,m)\mid T\big)
 =\alpha_t\frac{R_T-|\mathcal A(x)|}
 {|\mathcal S_{\min}(x)|}\,
 \mathbb I[m=m_v(T)]
 \label{eq:oct_bfs}
\end{equation}
for $x\preceq T$, $v\in\mathcal S_{\min}(x)$, and zero
otherwise. For $k<R_T$, with $n_\ell$ the
number of internal nodes of $T$ at depth $\ell$ and
$a_\ell$ the number of those already expanded, the
layered orders beginning with $\mathcal A(x)$ form a
fraction $1/\binom{n_\ell}{a_\ell}$ of all layered orders, so
\begin{equation}
 p_t(x\mid T)
 =\binom{R_T}{k}\kappa_t^k(1-\kappa_t)^{R_T-k}
 \Big/\binom{n_\ell}{a_\ell}
 \label{eq:app_oct_bfs_path}
\end{equation}
for consistent layered states with $k<R_T$, $p_t(T\mid T)=\kappa_t^{R_T}$ for $k=R_T$,
and zero otherwise. The model rates are
restricted to $\mathcal S_{\min}(x)$, and the training
loss is Eq.~\eqref{eq:oct_loss} with the sum taken over
$\mathcal S_{\min}(x)$ and $n_v(T)$ replaced by
$R_T(x)/|\mathcal S_{\min}(x)|$; inference restricts
expansions to $\mathcal S_{\min}(x)$ as detailed in
Appendix~\ref{app:oct_sampler}.

\subsection{Count-First Sampling under Uniform Time}
\label{app:oct_count_first}

The training measure samples $t\sim\mathcal U(0,1)$.
For a power schedule $\kappa_t=t^p$, writing $a=1/p$
gives
\begin{equation}
 \kappa\sim\mathrm{Beta}(a,1),
 \qquad
 K\mid\kappa,T
 \sim\mathrm{Binomial}(R_T,\kappa).
\end{equation}
By Beta--Binomial conjugacy,
\begin{equation}
 \begin{aligned}
 K\mid T
 &\sim\mathrm{BetaBinomial}(R_T,a,1),\\[-2pt]
 \kappa\mid K,T
 &\sim\mathrm{Beta}(K+a,R_T-K+1),
 \end{aligned}
 \label{eq:app_oct_beta_binomial}
\end{equation}
followed by $t=\kappa^{1/p}$. The order $\pi$ is
sampled independently of $(K,t)$: by sorting the keys
$\sigma_v$ of Eq.~\eqref{eq:oct_keys} for uniform
topological expansion, and, for breadth-first
expansion, by sorting $\operatorname{depth}(v)+U_v$ with
independent $U_v\sim\mathcal U(0,1)$. Its first $K$
elements define the expanded set.

The fully expanded count $K=R_T$ leaves no expandable
node and therefore has zero loss. To avoid processing
such samples, we draw $K$ from the Beta--Binomial law
conditioned on $K<R_T$, an event of probability
\begin{equation}
 \eta_T
 =\Pr(K<R_T\mid T)
 =1-\mathbb E[\kappa^{R_T}]
 =\frac{R_T}{R_T+a}.
 \label{eq:app_oct_terminal_weight}
\end{equation}
Multiplying the retained sample loss by $\eta_T$
preserves the original objective exactly.
Algorithm~\ref{alg:oct_train} summarizes one training step
for our default uniform topological expansion (with the
breadth-first variant replacing the key sort by
$\operatorname{depth}(v)+U_v$ and $\ell_{\mathrm{oct}}$
by the restricted loss of Appendix~\ref{app:oct_bfs}).

\begin{algorithm}[t]
\caption{Octree Flow Simulation-Free Training Step}
\label{alg:oct_train}
\small
\begin{algorithmic}[1]
\Require Target octree $T$ ($R_T=|\mathcal I(T)|$), condition $c$, $\kappa_t=t^p$ ($a=1/p$)
\State Draw $K\sim\mathrm{BetaBinomial}(R_T,a,1\mid K<R_T)$; set $\eta_T=\frac{R_T}{R_T+a}$
\State Draw $\kappa\sim\mathrm{Beta}(K+a,\,R_T-K+1)$ and set $t=\kappa^{1/p}$
\State Draw $\varepsilon_v\sim\mathrm{Exp}(n_v(T))$ for $v\in\mathcal I(T)$; set $\sigma_v=\sigma_{\mathrm{pa}(v)}+\varepsilon_v$ ($\sigma_{\mathrm{pa}(\mathrm{root})}=0$)
\State Sort $\mathcal I(T)$ by increasing $\sigma_v$ to obtain $\pi=(v_1,\ldots,v_{R_T})$
\State Form partial octree $x$ with expanded set $\mathcal A(x)=\{v_1,\ldots,v_K\}$
\State Predict $\{(\lambda_v^\theta,Q_v^\theta)\}_{v\in\mathcal S(x)}$ from frontier $\mathcal F(x)$ at $(t,c)$
\State \Return $\eta_T\,\ell_{\mathrm{oct}}(x,T;\theta)$ \Comment{Eq.~\eqref{eq:oct_loss}}
\end{algorithmic}
\end{algorithm}

\subsection{Factorized Loss and Optimal Factors}
\label{app:oct_factor_loss}

At a consistent state $x$ and $v\in\mathcal S(x)$, the
canonical conditional rates over patterns are
$\bar a_m=n_v(T)\,\mathbb I[m=m_v(T)]$, and the
canonical model rates are
$\bar b_m=\lambda_v^\theta Q_v^\theta(m)$.
Since $\sum_mQ_v^\theta(m)=1$,
\begin{equation}
 \sum_mD_\phi(\bar a_m,\bar b_m)
 =\lambda_v^\theta-n_v
 +n_v\log\frac{n_v}{\lambda_v^\theta}
 -n_v\log Q_v^\theta(m_v),
\end{equation}
with $n_v=n_v(T)$ and $m_v=m_v(T)$. Summing over
$v\in\mathcal S(x)$ gives Eq.~\eqref{eq:oct_loss} up to
terms independent of $\theta$. By
Eq.~\eqref{eq:pre_marginal}, the marginal canonical
rate of expanding $v$ with pattern $m$ is
$\mathbb E[n_v(T)\,\mathbb I[m=m_v(T)]\mid X_t=x,c]$;
summing over $m$ and normalizing gives the optimal
factors
\begin{equation}
 \begin{aligned}
 \lambda_v^*
 &=\mathbb E[n_v(T)\mid X_t=x,c],\\[-2pt]
 Q_v^*(m)
 &=\frac{
 \mathbb E[n_v(T)\mathbb I[m=m_v(T)]
 \mid X_t=x,c]}{\lambda_v^*},
 \end{aligned}
 \label{eq:oct_optimal}
\end{equation}
so $\lambda_v^*$ is the expected subtree size of $v$,
and $Q_v^*$ is a subtree-size-weighted pattern
posterior rather than the plain posterior of $m_v(T)$.
The weighting arises because a node fires at a rate
equal to its subtree size: patterns that open larger
subtrees are revealed sooner.

\subsection{Adaptive \texorpdfstring{$\tau$}{tau}-Leaping}
\label{app:oct_sampler}

Generation starts from the root-only state at $\tau=0$
and repeats the following step until
$\mathcal S(x)=\varnothing$. The network is evaluated
once at the current state and time $t$, where
$\tau=\tau_t$, giving $\lambda_v^\theta$ and
$Q_v^\theta$ for every $v\in\mathcal S(x)$; we write
$\Lambda=\sum_{v\in\mathcal S(x)}\lambda_v^\theta$. For
a hazard-time step $h$, every $v\in\mathcal S(x)$
expands independently with probability
$1-e^{-h\lambda_v^\theta}$ and draws its pattern from
$Q_v^\theta$. The instantiated children join the
frontier and become eligible in the next step, and
$\tau$ advances to $\tau+h$.

\paragraph{Expansions per Step.}
The expected number of expansions in a step,
$\sum_{v\in\mathcal S(x)}(1-e^{-h\lambda_v^\theta})$,
increases with $h$. We bound it by
$\min(|\mathcal S(x)|,M)$ for a maximal parallelism $M$,
setting $h_M=\infty$ when $|\mathcal S(x)|\le M$ and
finding the largest step $h_M$ that satisfies this bound
by bisection otherwise.

\paragraph{Missed Expansions.}
Conditioned on $T$, a node $v\in\mathcal S(x)$ expands at
time $s\in[0,h]$ within the step with density
$n_v(T)+O(h)$ and leaves children whose target subtrees
hold $n_v(T)-1$ further expansions. The conditional chain
would fire these at total rate $n_v(T)-1$ over the
remaining interval $h-s$, whereas the leap holds them until
the next step. Integrating over $s$ and averaging over the
posterior of $T$ gives, to leading order in $h$, the
expected number of missed expansions
\begin{equation}
 \tfrac12h^2\sum_{v\in\mathcal S(x)}
 \mathbb E\big[n_v(T)\big(n_v(T)-1\big)\mid X_t=x,c\big].
 \label{eq:app_oct_missed}
\end{equation}
The network predicts the first moment
$\lambda_v^*=\mathbb E[n_v(T)\mid X_t=x,c]$
(Eq.~\eqref{eq:oct_optimal}), so the sampler evaluates
each summand at $\lambda_v^\theta$ as
$\lambda_v^\theta(\lambda_v^\theta-1)_+$, dropping the
posterior variance $\operatorname{Var}(n_v(T)\mid X_t=x,c)$,
with $(z)_+=\max(z,0)$ since $n_v(T)\ge1$ counts $v$
itself. Dividing by the expected number $h\Lambda$ of
expansions in the step gives the missed-expansion ratio
$\tfrac12hW$ of Eq.~\eqref{eq:oct_missed}, with
$W=\Lambda^{-1}\sum_{v}\lambda_v^\theta
(\lambda_v^\theta-1)_+$. Keeping this ratio below a
tolerance $\epsilon$ gives $h\le2\epsilon/W$, which is
unbounded when $W=0$.

\paragraph{Step Size.}
The step is
\begin{equation}
 h=\min\big\{h_{\mathrm{grid}},\;2\epsilon/W,\;h_M\big\},
 \label{eq:app_oct_step}
\end{equation}
where $h_{\mathrm{grid}}=\tau_{t_{\mathrm{next}}}-\tau$ with
$t_{\mathrm{next}}=\min\{t_k:t_k>t\}$ for $t<t_{G-1}$ on a physical-time grid
$0=t_0<\cdots<t_G=1$, so that the network is evaluated at
least once per grid interval. Once $t\ge t_{G-1}$
(including $t=1$, $\tau=\infty$), we
set $h_{\mathrm{grid}}=\infty$, so the other two bounds
determine the step; an unbounded step expands every
expandable node. Algorithm~\ref{alg:oct_sample} summarizes the complete
sampler for uniform topological expansion.

\paragraph{Breadth-First Sampling.}
For the breadth-first variant (Appendix~\ref{app:oct_bfs}),
only minimal-depth frontier nodes $\mathcal S_{\min}(x)$
are eligible at each step, so deeper children wait until
the current depth completes and the missed-expansion
bound $2\epsilon/W$ is omitted ($h=\min\{h_{\mathrm{grid}},h_M\}$). In shallow layers
($\operatorname{depth}\le 2$), where few nodes carry
large rates $R_T(x)/|\mathcal S_{\min}(x)|$, we sample
one exponential waiting time
$\Delta\tau\sim\mathrm{Exp}(\Lambda)$ over
$\Lambda=\sum_{v\in\mathcal S_{\min}(x)}\lambda_v^\theta$
at a time, committing a single node proportional to
$\lambda_v^\theta$ if $\Delta\tau\le h_{\mathrm{grid}}$
and advancing $\tau$ by $\min\{\Delta\tau,h_{\mathrm{grid}}\}$, before
running Algorithm~\ref{alg:oct_sample} on
$\mathcal S_{\min}(x)$ for $\operatorname{depth}\ge 3$.

\begin{algorithm}[t]
\caption{Octree Flow Adaptive $\tau$-Leaping Sampler}
\label{alg:oct_sample}
\small
\begin{algorithmic}[1]
\Require Condition $c$, grid $0=t_0<\cdots<t_G=1$, $\epsilon$, parallelism $M$
\State Initialize root-only partial octree $x$, hazard time $\tau\gets 0$
\While{$\mathcal S(x)\ne\varnothing$}
    \State Set $t=\kappa^{-1}(1-e^{-\tau})$
    \State If $t<t_{G-1}$, set $t_{\mathrm{next}}\gets\min\{t_k:t_k>t\}$ and $h_{\mathrm{grid}}\gets\tau_{t_{\mathrm{next}}}-\tau$; else $h_{\mathrm{grid}}\gets\infty$
    \State Predict $\{(\lambda_v^\theta,Q_v^\theta)\}_{v\in\mathcal S(x)}$ from $\mathcal F(x)$ at $(t,c)$
    \State $\Lambda\gets\sum_{v\in\mathcal S(x)}\lambda_v^\theta$
    \State $W\gets\Lambda^{-1}\sum_{v\in\mathcal S(x)}\lambda_v^\theta(\lambda_v^\theta-1)_+$
    \If{$|\mathcal S(x)|\le M$}
        \State $h_M\gets\infty$
    \Else
        \State Solve $\sum_{v\in\mathcal S(x)}(1-e^{-h_M\lambda_v^\theta})=M$ for $h_M$ by bisection
    \EndIf
    \State $h\gets\min\{h_{\mathrm{grid}},\;2\epsilon/W,\;h_M\}$ \Comment{$2\epsilon/W=\infty$ if $W=0$}
    \ForAll{$v\in\mathcal S(x)$ \textbf{in parallel}}
        \State With prob.\ $1-e^{-h\lambda_v^\theta}$, draw $m_v\sim Q_v^\theta$ and expand $v$ (Eq.~\eqref{eq:oct_transition})
    \EndFor
    \State $\tau\gets\tau+h$
\EndWhile
\State \Return centers of depth-$L$ leaf cells in $\mathcal F(x)$ as vertices $V$
\end{algorithmic}
\end{algorithm}

\section{Additional Details of Edge Flow}
\label{app:edge}

\subsection{Directed Objective}
\label{app:edge_loss}

At a training state $E\subseteq E_1$ sampled from the
conditional path, let $y_{ij}=\mathbb I[\{i,j\}\in E_1]$
for every candidate pair $\{i,j\}\notin E$. Each pair has two directed
channels with canonical conditional rates
$\bar a_{i\to j}=\bar a_{j\to i}=\tfrac12 y_{ij}$ and
predictions $\bar b_{i\to j}=\lambda_i^\theta
Q_i^\theta(j)$. Since
$\sum_{j\in\mathcal N_i(E)}Q_i^\theta(j)=1$ whenever
$\mathcal N_i(E)\ne\varnothing$,
\begin{equation}
 \begin{aligned}
 &\sum_i\sum_{j\in\mathcal N_i(E)}
 D_\phi\big(\bar a_{i\to j},\bar b_{i\to j}\big)
 =\sum_{i:\mathcal N_i(E)\ne\varnothing}\lambda_i^\theta\\[-2pt]
 &\quad-\tfrac12\sum_{\{i,j\}\in E_1\setminus E}
 \log\big[\lambda_i^\theta Q_i^\theta(j)\,
 \lambda_j^\theta Q_j^\theta(i)\big]
 +\mathrm{const},
 \end{aligned}
\end{equation}
which is Eq.~\eqref{eq:edge_loss} up to the constant.
Below, $\delta_i$ denotes the number of target edges at
$i$ still missing from $E$; collecting
$\log\lambda_i^\theta$, which appears once for each of
them, shows that the rate term is
$\sum_{i:\mathcal N_i(E)\ne\varnothing}[\lambda_i^\theta
-\tfrac12\delta_i\log\lambda_i^\theta]$.

\paragraph{Relation to the Pair Objective.}
Let
$\ell_{\mathrm{pair}}=\sum_{\{i,j\}\notin E}
D_\phi(y_{ij},w_{ij}^\theta)$ be the
generalized-KL objective on the canonical pair rates,
and $s_{ij}=\lambda_i^\theta Q_i^\theta(j)/w_{ij}^\theta$.
Then
\begin{equation}
 \ell_{\mathrm{edge}}
 \simeq\ell_{\mathrm{pair}}
 +\!\sum_{\{i,j\}\in E_1\setminus E}\!\!
 \mathrm{KL}\big(
 (\tfrac12,\tfrac12)\,\|\,(s_{ij},1-s_{ij})
 \big).
 \label{eq:edge_balance}
\end{equation}
Write
$\mu_{ij}=\lambda_i^\theta Q_i^\theta(j)$ and
$\mu_{ji}=\lambda_j^\theta Q_j^\theta(i)$, so that
$w_{ij}^\theta=\mu_{ij}+\mu_{ji}$ and
$s_{ij}=\mu_{ij}/(\mu_{ij}+\mu_{ji})$. For
a non-target pair,
$D_\phi(0,\mu_{ij})+D_\phi(0,\mu_{ji})
=\mu_{ij}+\mu_{ji}=D_\phi(0,\mu_{ij}+\mu_{ji})$. For a
missing target edge,
\begin{equation}
 \begin{aligned}
 &D_\phi(\tfrac12,\mu_{ij})+D_\phi(\tfrac12,\mu_{ji})\\[-2pt]
 &\quad=\mu_{ij}+\mu_{ji}-1-\tfrac12\log(4\mu_{ij}\mu_{ji})\\[-2pt]
 &\quad=D_\phi(1,\mu_{ij}+\mu_{ji})\\[-2pt]
 &\qquad+\mathrm{KL}\big(
 (\tfrac12,\tfrac12)\,\|\,(s_{ij},1-s_{ij})\big),
 \end{aligned}
\end{equation}
since the KL term equals
$-\tfrac12\log\big(4\mu_{ij}\mu_{ji}/(\mu_{ij}+\mu_{ji})^2\big)$;
summing over candidate pairs gives Eq.~\eqref{eq:edge_balance}.

The KL term is nonnegative and vanishes at
$s_{ij}=\tfrac12$, and any nonnegative symmetric pair
rates admit this decomposition with $s_{ij}=\tfrac12$.
At any state $E$ with $p_t(E\mid V,c)>0$, minimizing
$\mathbb E[\ell_{\mathrm{edge}}\mid E_t=E,V,c]$ over
$(\lambda_i,Q_i)$ therefore minimizes
$\mathbb E[\ell_{\mathrm{pair}}\mid E_t=E,V,c]$, whose
minimizer is the marginal canonical rate
$w_{ij}^*=\Pr(\{i,j\}\in E_1\mid E_t=E,V,c)$ by
Eq.~\eqref{eq:pre_marginal}, and splits it equally,
which gives
\begin{equation}
 \begin{aligned}
 \lambda_i^*
 &=\tfrac12\,\mathbb E[\delta_i\mid E_t=E,V,c],\\[-2pt]
 Q_i^*(j)
 &=\frac{\Pr(\{i,j\}\in E_1\mid E_t=E,V,c)}{2\lambda_i^*}
 \end{aligned}
 \label{eq:edge_optimal}
\end{equation}
for $\lambda_i^*>0$ (and arbitrary on $\mathcal N_i(E)$
when $\lambda_i^*=0$).

With pointer logits $z_{ij}=q_i^\top k_j/\sqrt d$, the
gradient of the pointer term is
\begin{equation}
 \frac{\partial\ell_{\mathrm{edge}}}{\partial z_{ij}}
 =\tfrac12\big(\delta_i\,Q_i^\theta(j)-y_{ij}\big),
 \qquad j\in\mathcal N_i(E).
 \label{eq:edge_ptr_grad}
\end{equation}
Whenever $\delta_i>0$, the candidates of $i$ compete for
one unit of probability mass, pushing down each false
candidate ($y_{ij}=0$) in proportion to $\delta_i Q_i^\theta(j)$,
whereas when $\delta_i=0$ the pointer gradient vanishes
and $\lambda_i^\theta$ alone suppresses outgoing
proposals from $i$.

\subsection{Posterior Sampling with Temperature}
\label{app:edge_sampling}

Given $E_1$, target edges arrive independently, so the
event $e\notin E_t$ conditioned on $(E_1,E_t\setminus\{e\},V,c)$
depends only on $\mathbb I[e\in E_1]$, occurring with
probability $1-\kappa_t$ if $e\in E_1$ and $1$ otherwise.
For any candidate pair $e\notin E$ and any prior over $E_1$,
Bayes' rule gives the exact relation
\begin{equation}
 \operatorname{logit}w
 =\operatorname{logit}\tilde w+\log(1-\kappa_t),
 \label{eq:edge_pi}
\end{equation}
where $w=\Pr(e\in E_1\mid E_t=E,V,c)$ and $\tilde w$ is the
probability that $e\in E_1$ given $(V,c)$ and the states
of all other pairs at time $t$. At inference step $k$
(time $t_k$), we first form the clipped prediction
$\hat w_{ij}=\operatorname{clip}(w_{ij}^\theta,0,1)$. Keeping
the boundary values $\hat w_{ij}^\beta=\hat w_{ij}\in\{0,1\}$ fixed,
for $\hat w_{ij}\in(0,1)$ we set
$u_{ij}=\operatorname{logit}\hat w_{ij}-\log(1-\kappa_{t_k})$
and evaluate $\hat w_{ij}^\beta=\sigma(u_{ij}/\beta+\log(1-\kappa_{t_k}))\in(0,1)$
in Eq.~\eqref{eq:edge_sample} at temperature $\beta>0$,
where $\sigma(u)=(1+e^{-u})^{-1}$ is the logistic sigmoid.
The factor $\gamma_k$ is unchanged, so the temperature
changes which edges are preferred but not when they are
decided. Algorithm~\ref{alg:edge} summarizes simulation-free
training and posterior sampling for Edge Flow.

\begin{algorithm}[t]
\caption{Edge Flow Training and Posterior Sampling}
\label{alg:edge}
\small
\begin{algorithmic}[1]
\Statex \textbf{Simulation-Free Training Step} on $(V,E_1,c)$:
\State Draw $t\sim\mathcal U(0,1)$; keep each $e\in E_1$ independently with prob.\ $\kappa_t$ to form $E$
\State Predict $\{\lambda_i^\theta,q_i,k_i\}_{i=1}^N$ on $(V,E,t,c)$; compute $Q_i^\theta$ over $\mathcal N_i(E)$
\State \Return $\ell_{\mathrm{edge}}(E,E_1;\theta)$ \Comment{Eq.~\eqref{eq:edge_loss}}
\vspace{3pt}
\makeatletter\renewcommand{\theHALG@line}{\thealgorithm.s.\arabic{ALG@line}}\setcounter{ALG@line}{0}\makeatother
\Statex \textbf{Posterior Sampler} on $(V,c)$ with grid $0=t_0<\cdots<t_S=1$, temp.\ $\beta$:
\State Initialize $E\gets\varnothing$
\For{$k=0,\ldots,S-1$}
    \State Predict $\{\lambda_i^\theta,q_i,k_i\}_{i=1}^N$ on $(V,E,t_k,c)$ and compute $Q_i^\theta$ over $\mathcal N_i(E)$
    \State Set $\gamma_k=(\kappa_{t_{k+1}}-\kappa_{t_k})/(1-\kappa_{t_k})$ \Comment{$\gamma_{S-1}=1$}
    \ForAll{$\{i,j\}\notin E$ \textbf{in parallel}}
        \State $\hat w_{ij}\gets\operatorname{clip}\big(\lambda_i^\theta Q_i^\theta(j)+\lambda_j^\theta Q_j^\theta(i),\,0,\,1\big)$
        \If{$\hat w_{ij}\in(0,1)$}
            \State $u_{ij}\gets\operatorname{logit}\hat w_{ij}-\log(1-\kappa_{t_k})$
            \State $\hat w_{ij}^\beta\gets\sigma\big(u_{ij}/\beta+\log(1-\kappa_{t_k})\big)$
        \Else
            \State $\hat w_{ij}^\beta\gets\hat w_{ij}$
        \EndIf
        \State Insert $\{i,j\}$ into $E$ independently with probability $\gamma_k\hat w_{ij}^\beta$
    \EndFor
\EndFor
\State \Return $E$ and recover faces as the 3-cliques of $E$
\end{algorithmic}
\end{algorithm}

\section{Implementation and Training Details}
\label{app:impl}

\paragraph{\methodname Architecture, Training, and Sampling.}
Point-cloud models encode $40{,}960$ surface points with normals (augmented with position and normal noise during training) into $2{,}048$ condition tokens via a VecSet encoder~\citep{zhang20233dshape2vecset} trained jointly with the Transformer (the 16K-point results in Tab.~\ref{tab:main} feed the same models $16{,}384$ points at inference), whereas image-conditioned models use a frozen DINOv2-L~\citep{oquab2024dinov2} encoder on $518\times 518$ inputs.
Octree Flow uses a $24$-layer, $768$-width Transformer ($\sim\!370\text{M}$ parameters) for point clouds and a $25$-layer, $1024$-width Transformer ($\sim\!670\text{M}$ parameters) for images, trained for $500\text{K}$ steps under $\kappa_t=t^3$ via Algorithm~\ref{alg:oct_train}; Edge Flow uses a $25$-layer, $1024$-width Graph Transformer ($\sim\!700\text{M}$ parameters, pointer query/key dimension $d=128$) trained for $200\text{K}$ steps under $\kappa_t=t$ via Algorithm~\ref{alg:edge}.
All models are trained on $\sim\!1.17\text{M}$ filtered Objaverse-XL~\citep{deitke2023objaversexl} meshes ($<8\text{K}$ vertices, normalized to $[-1,1]^3$, $L=10$) using AdamW with learning rate $10^{-4}$, global batch size $128$, $10\%$ condition dropout for classifier-free guidance (CFG), and EMA weights, on $8$ GPUs for point-cloud models and $16$ GPUs for image-conditioned models.
At inference, Octree Flow runs Algorithm~\ref{alg:oct_sample} ($G=50$ uniform physical-time grid, $\epsilon=0.05$, $M=128$, sampling $m_v$ at pattern temperature $0.1$, with CFG $f_{\mathrm{cond}}+w(f_{\mathrm{cond}}-f_{\mathrm{uncond}})$ on log-intensities and pattern logits using $w=0$ on Toys4K/Objaverse, $w=1.0$ on Dense Mesh, and $w=2.0$ for images), and Edge Flow runs Algorithm~\ref{alg:edge} ($S=50$ uniform steps, $\beta=0.3$, without CFG).

\paragraph{Controlled \nexus Reimplementation.}
\nexus matches the point-cloud \methodname setup in training data, $1024^3$ ($L=10$) quantization grid, jointly trained VecSet conditioner, Transformer backbones ($24\times 768$ for vertices; $25\times 1024$ for connectivity), optimizer, batch size, and step budgets ($500\text{K}$ and $200\text{K}$ steps).
For vertex generation, the cascaded octree model is trained with continuous flow matching over relaxed $8$-child occupancies by uniformly sampling an octree depth $\ell\sim\mathrm{Uniform}\{0,\ldots,L-1\}$ and flow time $t\sim\mathcal U(0,1)$ on the ground-truth depth-$\ell$ frontier, and sampled coarse-to-fine from $\ell=0$ to $L-1$ with $20$ steps per depth ($200$ NFEs in total).
For connectivity generation, we first train a topology autoencoder strictly following the architecture and balanced spacetime-interval objective of Nexus~\citep{wang2026nexus}; conditioned on $V$ and $c$, Latent Flow is trained with continuous flow matching under $t\sim\mathcal U(0,1)$ to predict per-vertex topology latents and sampled in $50$ steps (matching Edge Flow's budget) before decoding edges and faces.
Released baselines use their official checkpoints and input interfaces; in particular, LATO.2~\citep{long2026lato2} takes a rendered image and reference sparse voxels in Tab.~\ref{tab:main} ($^\ddagger$) and ground-truth vertices in Tab.~\ref{tab:gt_vertices}.

\section{Evaluation Benchmarks and Metrics}
\label{app:benchmarks}

\begin{figure*}[p]
    \centering
    \begin{subfigure}{0.64\textwidth}
        \centering
        \includegraphics[width=\linewidth]{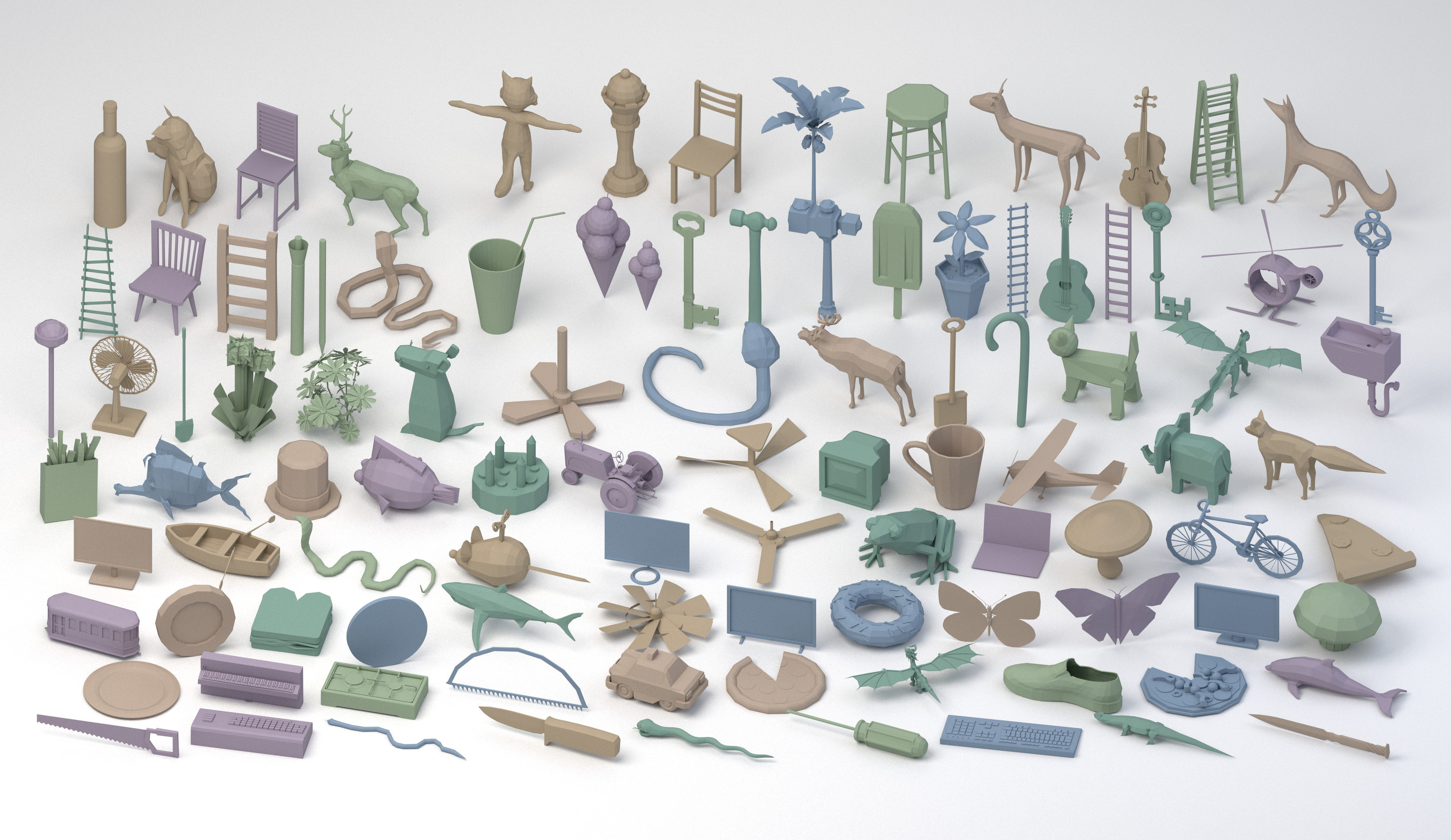}
        \vspace{-1.5mm}
        \caption{\textbf{Toys4K}}
        \label{fig:supp_toys4k_overview}
    \end{subfigure}

    \vspace{0.8mm}
    \begin{subfigure}{0.64\textwidth}
        \centering
        \includegraphics[width=\linewidth]{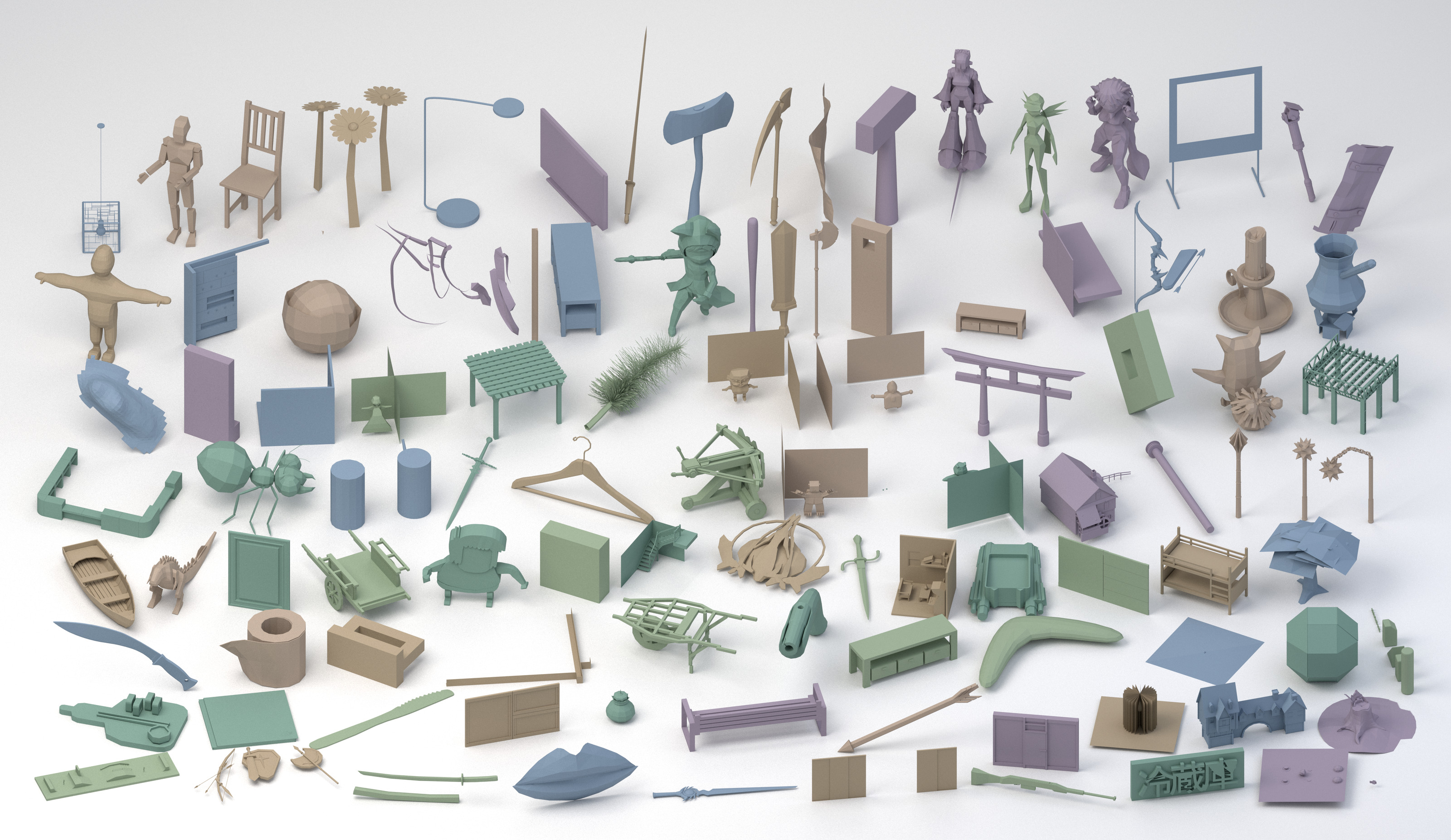}
        \vspace{-1.5mm}
        \caption{\textbf{Objaverse}}
        \label{fig:supp_objaverse_overview}
    \end{subfigure}

    \vspace{0.8mm}
    \begin{subfigure}{0.64\textwidth}
        \centering
        \includegraphics[width=\linewidth]{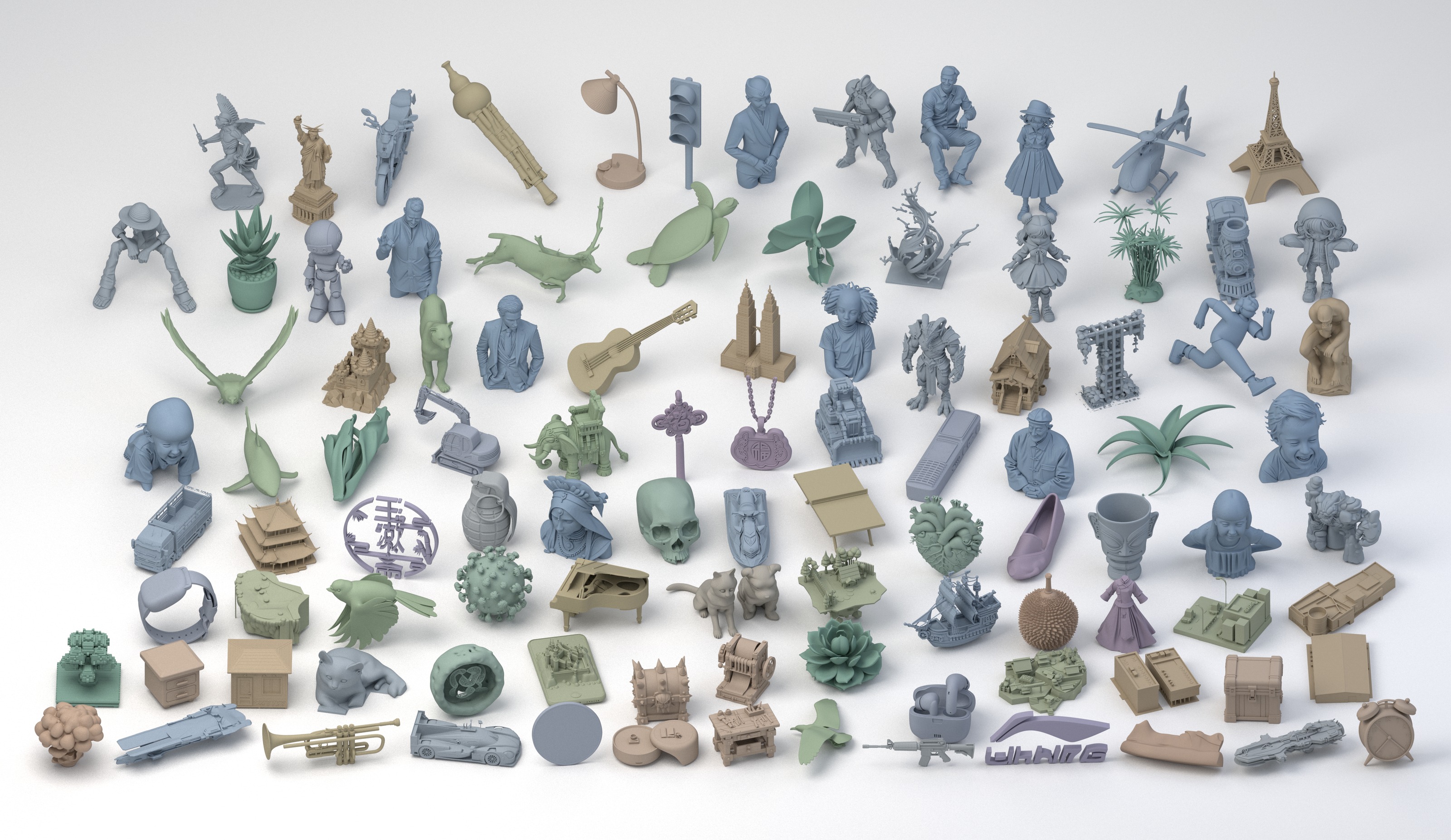}
        \vspace{-1.5mm}
        \caption{\textbf{Dense Mesh}}
        \label{fig:supp_dense_overview}
    \end{subfigure}
    \caption{\textbf{Overviews of $100$ randomly selected meshes} from the held-out (a) Toys4K, (b) Objaverse, and (c) Dense Mesh evaluation sets.}
    \label{fig:supp_dataset_overviews}
\end{figure*}

\paragraph{Benchmarks.}
We evaluate on two held-out artistic mesh benchmarks strictly disjoint from the training set, retaining shapes with $\le 4{,}000$ faces ($64$--$8{,}136$ vertices) to remain within the sequence-length budgets of autoregressive baselines: \textbf{Toys4K} ($1{,}068$ shapes; all qualifying meshes in the dataset) and \textbf{Objaverse} ($500$ randomly sampled qualifying shapes).
To evaluate dense-to-artistic mesh conversion without face-count filtering, we additionally test on \textbf{Dense Mesh}, comprising $222$ generated dense meshes~\citep{xiang2025trellis2,lai2026lattice,jia2025ultrashape} ($\sim\!500\text{K}$ triangles per shape); as BPT~\citep{weng2025bpt} fails to produce valid meshes on $27$ shapes (whereas \methodname succeeds on all $222$), Tab.~\ref{tab:main} evaluates all methods on the common $195$-shape subset.
Fig.~\ref{fig:supp_dataset_overviews} shows $100$ randomly selected meshes from each benchmark.

\paragraph{Evaluation Metrics.}
All predicted and reference meshes are evaluated in the normalized $[-1,1]^3$ coordinate frame.
Given a predicted triangle mesh $\mathcal M$ and reference (ground-truth) mesh $\mathcal M_{\mathrm{gt}}$, let
$d(p,\mathcal M)=\min_{f\in\mathcal M}\min_{q\in f}\|p-q\|_2$
be the exact Euclidean point-to-triangle distance from $p\in\mathbb R^3$ to $\mathcal M$, and let $n_{\mathcal M}(p)$ denote the unit geometric normal of a closest nondegenerate triangle $f\in\mathcal M$ to $p$.
We draw $N_{\mathrm{s}}=10^5$ area-uniform surface points $S_{\mathrm{pred}}$ on $\mathcal M$ and $S_{\mathrm{gt}}$ on $\mathcal M_{\mathrm{gt}}$ (with source-triangle unit normals $n(p)$ and $n(q)$), and compute:
\begin{align}
 \mathrm{CD}
 &=\frac{1}{2N_{\mathrm{s}}}\!\left(
 \sum_{p\in S_{\mathrm{pred}}}\!d(p,\mathcal M_{\mathrm{gt}})
 +\!\sum_{q\in S_{\mathrm{gt}}}\!d(q,\mathcal M)\right),\\
 \mathrm{HD}
 &=\max\!\left(
 \max_{p\in S_{\mathrm{pred}}}d(p,\mathcal M_{\mathrm{gt}}),\;
 \max_{q\in S_{\mathrm{gt}}}d(q,\mathcal M)\right),\\
 \begin{split}
 \mathrm{NC}
 &=\frac{1}{2N_{\mathrm{s}}}\!\sum_{p\in S_{\mathrm{pred}}}\!
 \big|\langle n(p),n_{\mathcal M_{\mathrm{gt}}}(p)\rangle\big|\\[-2pt]
 &\quad+\frac{1}{2N_{\mathrm{s}}}\!\sum_{q\in S_{\mathrm{gt}}}\!
 \big|\langle n(q),n_{\mathcal M}(q)\rangle\big|,
 \end{split}
\end{align}
reporting $\mathrm{CD}\times 10^3$ and $\mathrm{HD}\times 10^2$.
Using absolute inner products evaluates geometric tangent-plane alignment independently of face winding orientation.
When evaluating connectivity generation on ground-truth vertices $V$ (Tabs.~\ref{tab:gt_vertices} and~\ref{tab:ablation_edge}), let $E$ and $E_{\mathrm{gt}}$ be the predicted and ground-truth sets of undirected edges over $V$; undirected \textbf{Edge F1} (\%) is
\begin{equation}
 \mathrm{Edge\ F1}(E,E_{\mathrm{gt}})
 =100\times\frac{2\,|E\cap E_{\mathrm{gt}}|}{|E|+|E_{\mathrm{gt}}|}.
\end{equation}

\section{Additional Qualitative Results}
\label{app:qualitative}

\begin{figure*}[p]
    \centering
    \includegraphics[width=\textwidth]{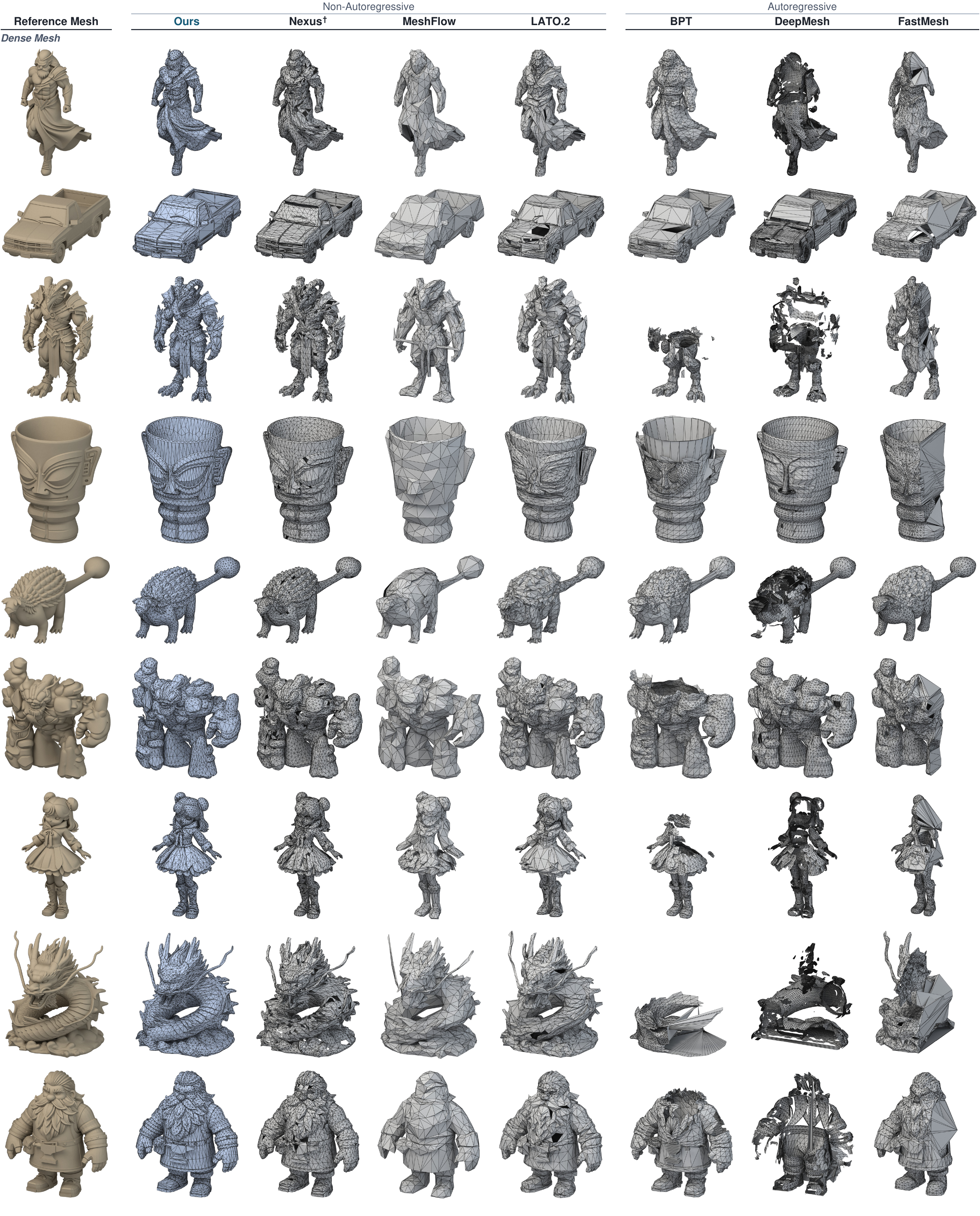}
    \caption{\textbf{Additional point-cloud-conditioned mesh generation} on Dense Mesh ($^\dagger$controlled reimplementation).}
    \label{fig:supp_dense_qualitative}
\end{figure*}

\begin{figure*}[p]
    \centering
    \includegraphics[width=\textwidth]{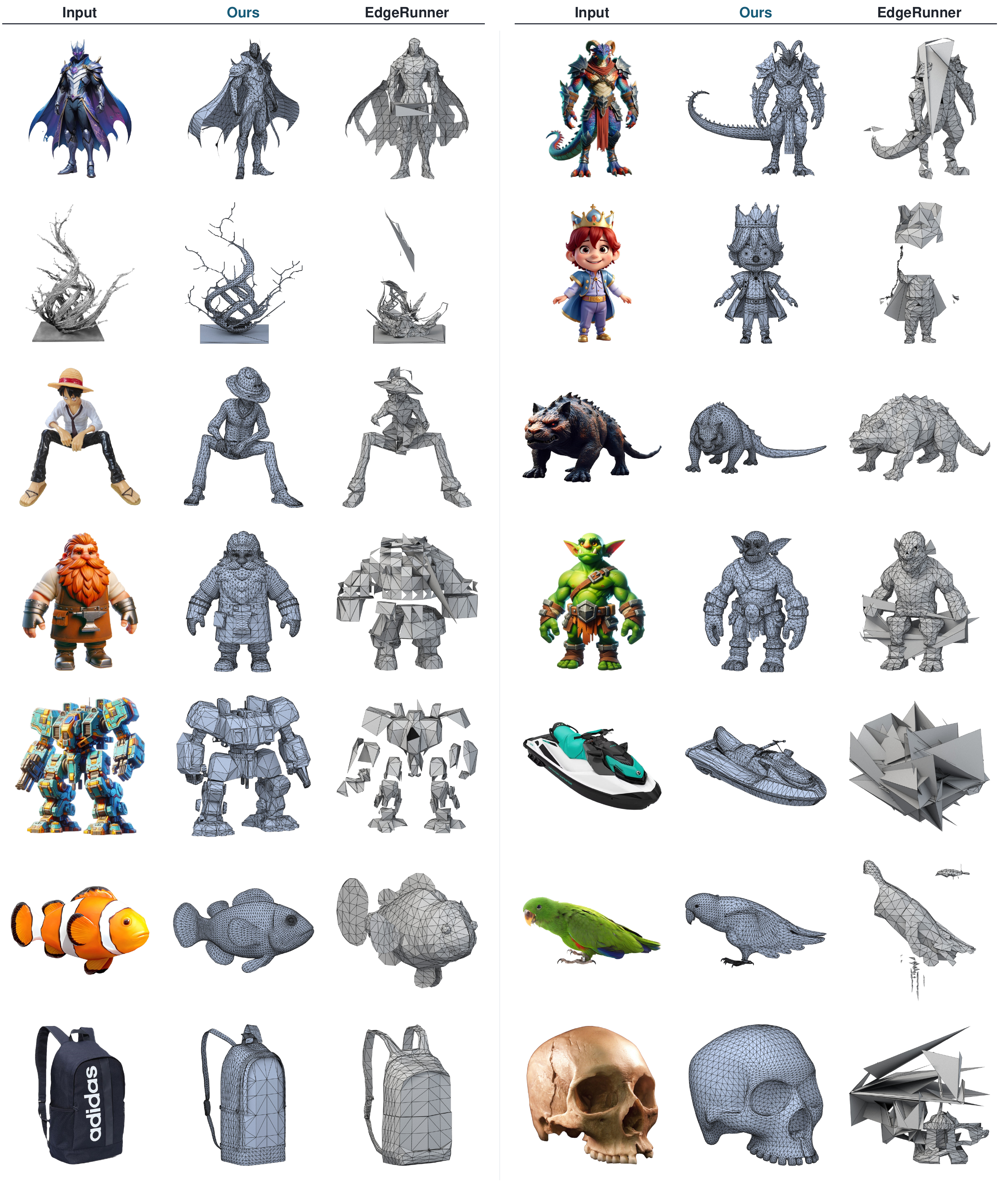}
    \caption{\textbf{Additional single-image-to-mesh generation} vs.\ EdgeRunner~\citep{tang2025edgerunner}.}
    \label{fig:supp_image_to_mesh}
\end{figure*}

Fig.~\ref{fig:supp_dense_qualitative} presents additional point-cloud-conditioned comparisons on Dense Mesh, and Fig.~\ref{fig:supp_image_to_mesh} presents additional single-image-to-mesh comparisons with EdgeRunner~\citep{tang2025edgerunner}.

\end{document}